\documentclass[lettersize,journal]{IEEEtran}
\usepackage{array}
\usepackage{multicol}
\usepackage[caption=false,font=normalsize,labelfont=sf,textfont=sf]{subfig}
\usepackage{textcomp}
\usepackage{stfloats}
\usepackage{url}
\usepackage{verbatim}
\usepackage{graphicx}
\usepackage{tikz}
\usepackage{graphicx}
\usepackage{cite}
\usepackage{multirow}
\usepackage{makecell}

\usepackage[ruled,linesnumbered]{algorithm2e}
\usepackage{amsmath}
\usepackage{bm}
\usepackage{color}
\usepackage{graphicx}
\usepackage{amssymb}
\usepackage{url}
\usepackage{booktabs}
\usepackage{multirow}
\usepackage[normalem]{ulem}
\usepackage{ragged2e}
\usepackage{soul}
\usepackage{xcolor}
\usepackage{ulem}% colors
\usepackage{colortbl}
\usepackage{booktabs} 
\usepackage[colorlinks,linkcolor=black,citecolor=black]{hyperref}
\usepackage{doi} % 提供\doi命令

\useunder{\uline}{\ul}{}

\usepackage{longtable}
\usepackage{relsize}
\usepackage{comment}
\usepackage{diagbox}

\begin{document}
%Synthetic 
%\title{Genetic Oversampling Method Based on Evolutionary Multi-task Learning}
\title{Evo-TFS: Evolutionary Time-Frequency Domain-Based Synthetic
Minority Oversampling Approach to Imbalanced Time Series Classification}

\author{Wenbin Pei*, Ruohao Dai*, Bing Xue,~\IEEEmembership{Fellow,~IEEE,}	Mengjie Zhang,~\IEEEmembership{Fellow,~IEEE}, Qiang Zhang,
Yiu-Ming Cheung,~\IEEEmembership{Fellow,~IEEE}}

\markboth{IEEE Transactions on Evolutionary Computation}%
{Shell \MakeLowercase{\textit{et al.}}: A Sample Article Using IEEEtran.cls for IEEE Journals}

% \IEEEpubid{0000--0000/00\$00.00~\copyright~2021 IEEE}
% Remember, if you use this you must call \IEEEpubidadjcol in the second
% column for its text to clear the IEEEpubid mark.

\maketitle

\begin{abstract}
Time series classification is a fundamental machine learning task with broad real-world applications. Although many deep learning methods have proven effective in learning time-series data for classification, they were originally developed under the assumption of balanced data distributions. Once data distribution is uneven, these methods tend to ignore the minority class that is typically of higher practical significance. %yet class imbalance remains a major challenge in real-world applications. This imbalance often results in biased classifiers that underperform on minority classes, which are typically of higher practical significance.   
Oversampling methods have been designed to address this by generating minority-class samples, but their reliance on linear interpolation often hampers the preservation of temporal dynamics and the generation of diverse samples. Therefore, in this paper, we propose Evo-TFS, a novel evolutionary oversampling method that integrates both time- and frequency-domain characteristics. In Evo-TFS, strongly typed genetic programming is employed to evolve diverse, high-quality time series, guided by a fitness function that incorporates both time-domain and frequency-domain characteristics.
Experiments conducted on imbalanced time series datasets demonstrate that Evo-TFS outperforms existing oversampling methods, significantly enhancing the performance of time-domain and frequency-domain classifiers. %\textcolor{red}{Further analysis indicates that the samples generated by Evo-TFS enable different classes in the dataset to have stronger density consistency.}
%Further analysis reveals that Evo-TFS produces more uniformly distributed samples. %, contributing to improved model generalization.
\end{abstract}

\begin{IEEEkeywords}
Imbalanced Time Series Classification, Oversampling, Genetic Programming, and Time-Frequency Domain.
\end{IEEEkeywords}

\section{Introduction}
\begin{figure}
  \centering
  \includegraphics[width=0.45\textwidth]{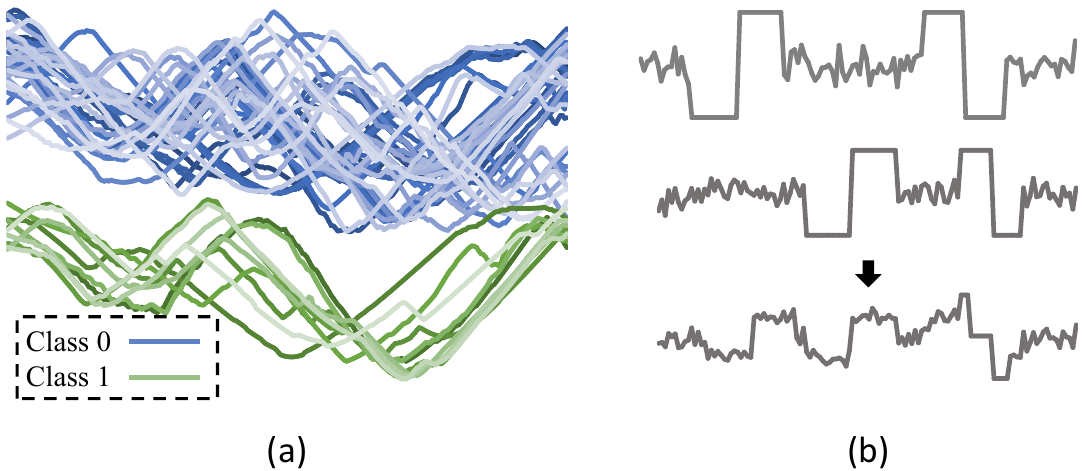}
  \caption{(a) Visualization of imbalanced time series datasets, in which the sample size of one type is significantly smaller; (b) Generating samples using minority class samples by interpolation methods.}
  \label{fig:intro}
\end{figure}

Time series classification (TSC) has become essential across a wide range of real-world domains, including financial risk assessment \cite{LI2025104025}, clinical diagnosis \cite{2020Utilizing}, industrial predictive maintenance \cite{jiang2019gan}, and cybersecurity \cite{sayegh2024enhanced}. Deep learning models, such as convolutional neural networks (CNNs) and long short-term memory networks (LSTMs) \cite{zhao2017convolutional,karim2019multivariate,mohammadi2024deep}, have demonstrated their strong effectiveness in capturing temporal dependencies and achieving high performance on complex time-series data. \textcolor{black}{However, deep learning models were initially developed under the assumption of balanced time-series data \cite{johnson2019survey}.} Unfortunately, in many real-world scenarios, such as medical diagnosis \cite{2020Utilizing} and fraud detection \cite{wang2023fraud}, time-series data is typically imbalanced, with certain, often important, classes having significantly fewer samples than others, as shown in Fig.~\ref{fig:intro} (a). \textcolor{black}{Learning from imbalanced time-series data poses a significant challenge for existing deep learning methods in TSC~\cite{ghosh2024class}}. %多数类梯度占据主导
The predominance of the majority-class samples leads these models to optimize disproportionately for the majority class, often at the expense of the performance on the minority class.

%This issue is particularly pronounced in fields like medical diagnosis (e.g., rare disease detection) and fraud detection (e.g., credit card fraud).

%Although deep learning models have shown great promise in TSC tasks, they still face significant challenges when handling imbalanced data. 

An effective solution to addressing class imbalance involves rebalancing the data distribution through sampling techniques. %is to apply oversampling at the data level, artificially rebalancing the distribution by augmenting minority class instances. 
Among them, one widely used technique is interpolation-based oversampling \cite{he2009learning,chawla2002smote,he2008adasyn,fernandez2018smote}, as illustrated in Fig.~\ref{fig:intro} (b). Note that time series inherently exhibit temporal order, whereas these interpolation methods usually focus solely on point-to-point value estimation. This disrupts the original temporal patterns and rhythms, causing synthetic samples to fail to accurately reflect the true temporal dynamics of the data \cite{ijcai2021p631}. To address this limitation, time-based synthetic minority over-sampling technique (T-SMOTE) \cite{zhao2022t} has been proposed, which incorporates temporal information during the data generation process to preserve time dependencies. Nevertheless, although T-SMOTE emphasizes time-domain characteristics, it tends to neglect frequency-domain characteristics that are critical in numerous real-world applications, including medical diagnostics, speech processing, and sensor data analysis. Therefore, an alternative approach is desired, which is expected to effectively integrate both time-domain and frequency-domain characteristics while ensuring the diversity and structural integrity of the generated samples\cite{ijcai2021p631}. 

%Therefore, there is a pressing need for an alternative approach that can effectively leverage both time-domain and frequency-domain features, while ensuring the diversity and structural integrity of the generated samples.
%To address these challenges, it is essential to explore oversampling strategies that go beyond simple interpolation, offering greater flexibility and expressive power.  

%Genetic programming (GP) \cite{poli2008field,langdon2013foundations,ahvanooey2019survey}, an evolutionary algorithm inspired by natural selection, offers a promising approach to developing an oversampling method to generate time series for TSC. Notably, GP evolves programs through evaluation, selection, and reproduction, enabling the automatic discovery of solutions without the need for explicit programming. 
Genetic programming (GP) \cite{poli2008field,langdon2013foundations,ahvanooey2019survey} is an evolutionary algorithm inspired by natural selection. It evolves programs through evaluation, selection, and reproduction, enabling the automatic discovery of solutions without the need for explicit programming. 
Typically, a GP individual is represented as a tree structure comprising a diverse set of configurable function nodes and terminal nodes. This gives GP great potential to flexibly construct a wide range of mathematical expressions, each of which can be regarded as a generated sample \cite{10793073}. 
By learning complex transformations from existing data, GP is capable of generating synthetic time series that capture intricate temporal patterns.
%-based structure allows for flexible program representation, while a rich set of customizable functions ensures that GP can construct a wide range of mathematical expressions suited to the task.  By learning complex transformations from existing data, GP is capable of generating synthetic time series that capture intricate temporal patterns.
%Moreover, both time-domain and frequency-domain information can be incorporated into the fitness function, guiding the evolution of candidate solutions toward preserving the temporal dependencies and spectral characteristics inherent in the original data.  
Moreover, both time-domain and frequency-domain information can be incorporated into the fitness function, guiding the evolution toward better solutions that maintain the temporal and spectral integrity of the original data. %This joint consideration of both time and frequency components enhances the diversity and representativeness of the generated time series for TSC tasks.

However, the potential of using GP to generate time series remains unexplored. Therefore, in this paper, we propose a novel evolutionary time-frequency-based synthetic minority over-sampling method (Evo-TFS) for imbalanced TSC tasks. %Evo-TFS uses multi-population GP to generate diverse synthetic time series for the minority class while maintaining both diversity and fidelity. 
Evo-TFS uses GP to generate diverse and high-quality time series data for the minority class. In Evo-TFS, each GP individual represents a time-series sample to be generated. To enhance the realism of the generated time series, both time-domain and frequency-domain information are incorporated into the fitness function. This ensures that the synthesized data maintains not only temporal dependencies but also spectral characteristics. The main contributions of Evo-TFS are introduced as follows: 

\begin{itemize}

%\item We introduce a knowledge transfer method that enhances the efficiency of GP evolution by transferring knowledge between GP sub-populations across different tasks.

\item We propose a novel individual representation, with terminal and function sets specifically designed for time series generation. This leverages GP to synthesize time series automatically and effectively. 

\item We design a new fitness function to evaluate the goodness of each individual. Both the time-domain and frequency-domain information are incorporated into the proposed fitness function during the sample generation process. This ensures that the synthetic samples have better time series characteristics.

\item We propose a new GP-based oversampling method specifically designed for generating time series in imbalanced TSC. Experimental results demonstrate that Evo-TFS outperforms existing oversampling techniques in generating time series, significantly enhancing the performance of time-domain and frequency-domain classifiers. %Further analysis reveals that Evo-TFS produces more uniformly distributed samples. %, contributing to improved model generalization.%Multi-population GP was used to generate synthetic samples, which increased the diversity of synthetic samples.

%\item 

\end{itemize}

\section{Background and related works}

%\subsection{Time Series Classification}
%这部分介绍TSC
\subsection{Imbalanced Time Series Classification}

In general, two widely used methods to address the class imbalance issue are cost-sensitive learning \cite{araf2024cost,10068793} and sampling techniques \cite{mohammed2020machine,bej2021loras}. %Class imbalance is a common issue in TSC tasks.

Cost-sensitive learning encourages learning models to treat different errors differently by explicitly considering the varying costs associated with different types of classification errors \cite{pei2023survey}. Unlike traditional learning algorithms, cost-sensitive methods aim to minimize the total misclassification cost rather than just the overall error rate \cite{elkan2001foundations}. Cost-sensitive learning can enhance model performance in imbalanced scenarios by modifying the training data distribution or incorporating cost matrices into the loss function. To date, many traditional classification algorithms have been adapted into cost-sensitive versions for use in imbalanced TSC tasks. For example, Cost-sensitive CNN modifies the loss function to adaptively train a cost-sensitive model for solving imbalanced TSC problems\cite{geng2019cost}. A robust time series
anomaly detection framework (RobustTAD) \cite{gao2020robusttad} introduces both label-based and value-based weights into the loss function to adjust the importance of class labels and the contribution of each sample’s neighborhood. However, cost-sensitive learning methods have several limitations. First, in real-world applications, humans, even domain experts, may find it difficult to accurately quantify cost matrices. Moreover, learning models may become more complicated when incorporating cost information and fail to generalize well once the cost information shifts or is incorrectly identified. 

%In general, it incorporates a cost matrix to assign different penalty weights to misclassifications of different classes, thereby enhancing the model's ability to identify minority classes. 

Different from cost-sensitive learning, sampling techniques aim to rebalance the training data to improve model performance in imbalanced scenarios. Random undersampling is often criticized for discarding potentially important samples, while random oversampling may lead to overfitting because of simply duplicating minority-class samples\cite{oksuz2020imbalance}. To mitigate these issues, the synthetic minority over-sampling technique (SMOTE) generates synthetic samples by linearly interpolating between minority-class samples in the feature space \cite{chawla2002smote}. Its variants, such as the adaptive synthetic sampling approach for imbalanced learning (ADASYN) method\cite{he2008adasyn} and Borderline-SMOTE\cite{han2005borderline}, further improve data quality of the generated samples by focusing on generating synthetic samples in specific regions of the minority class. For time series data, T-SMOTE \cite{zhao2022t} has been specifically designed to preserve the temporal continuity and dependency during the synthetic sample generation process, making it more suitable for imbalanced TSC tasks. However, T-SMOTE relies on interpolation between existing minority-class samples, which may make it less effective once the original sequences contain non-linear or non-stationary patterns. Moreover, T-SMOTE generates new samples based on local neighborhood information, which limits the diversity of the generated samples and may increase the risk of model overfitting.

\subsection{Genetic Programming}

\begin{figure}[h]
    \centering
	\includegraphics[
 width=0.42\textwidth]{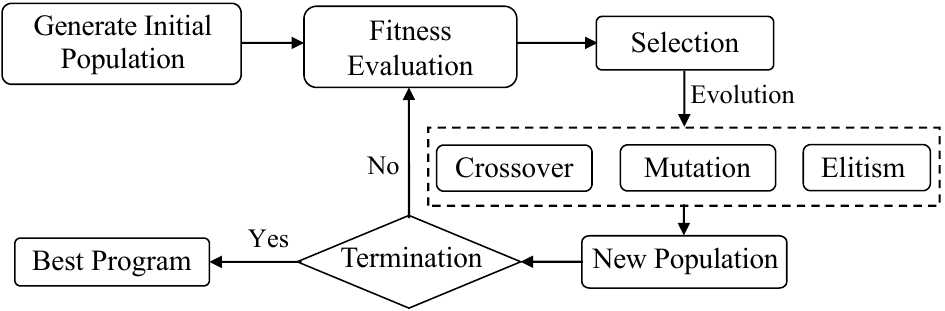}
	\caption{The flowchart of GP.} %(c) GP-based approaches with transfer optimization transfer information between populations with similar individuals in the feature space.} 
	\label{fig:GP}
\end{figure}

GP is an evolutionary algorithm, which has been widely used in symbolic regression, classification, prediction, and data generation tasks\cite{zhang2021surrogate,10793073,wang2025semantics}. The detailed GP process is illustrated in Fig.~\ref{fig:GP}. %For data generation, GP evolves a population of individuals through genetic operators, where each individual can be viewed as a candidate data sample. The goal is to generate optimal data, and the detailed process is illustrated in Fig.~\ref{fig:GP}.

Tree-based GP is the most widely used version. It requires only the prior definition of a function set, a terminal set, and a fitness function. Strongly-Typed genetic programming (STGP) improves the standard GP\cite{fonseca2023comparing,montana1995strongly} by introducing data type constraints, specifying the output types of terminals, and enforcing matching input and output types for functions. This enables STGP to handle multiple data types more effectively while reducing the search space.

In STGP, type-constrained initialization is used to initialize the population. Type-preserving crossover and type-constrained mutation serve as the main genetic operators, aiming to generate hopefully better offspring with improved fitness while adhering to type constraints.
\begin{itemize}

\item \textbf{Type-constrained Initialization:} A tree is constructed from the root by selecting functions whose return types match the required type, and recursively selects their arguments with functions or terminals of matching types. This ensures that all individuals in the initial population follow the type constraints and are therefore valid.

\item \textbf{Type-preserving Crossover:} A subtree is randomly selected from parent $P_1$, and another subtree with the same output type is randomly selected from parent tree $P_2$. These subtrees are then swapped to produce two offspring, ensuring type consistency and avoiding type conflicts.

\item \textbf{Type-constrained Mutation:} A subtree is randomly selected from a parent tree, and a newly generated tree with the same output type is randomly created to replace the subtree.

\end{itemize}

Note that a GP individual consists of configurable function nodes and terminal nodes. This gives GP great potential to flexibly construct a wide range of mathematical expressions, each of which can be regarded as a generated sample. By learning complex transformations from existing data, GP is capable of generating synthetic time series that capture intricate temporal patterns. However, to the best of our knowledge, there is no existing literature that explores the use of GP for generating dynamic time series data. %However, GP has a natural advantage in generating new samples. 
Therefore, this study aims to investigate how GP can be used to generate a high-quality set of time series data for imbalanced TSC tasks.

\section{The Proposed Method}
This section presents Evo-TFS, a time-frequency domain-based evolutionary oversampling approach to imbalanced TSC tasks. First, the overall framework of Evo-TFS is introduced. Subsequently, we detail the representation of the GP program specifically designed for generating time series data. Following this, the time-frequency domain-based fitness function is elaborated. Finally, the overall process of Evo-TFS is comprehensively explained.

%, along with the fitness function evaluation method based on the time-frequency domain. Finally, the evolutionary process of Evo-TFS is explained. 
%Subsequently, the GP program representation specifically designed for time series data is described. Following this, the time-frequency domain-based fitness function evaluation method is elaborated. Finally, the evolutionary process of the GP implementation is comprehensively explained.
%an evolutionary time-frequency-based synthetic minority over-samplin approach to imbalanced TSC tasks. 
\subsection{The Framework of Evo-TFS}

\begin{figure*}
\centering
  \includegraphics[width=1.0\textwidth]{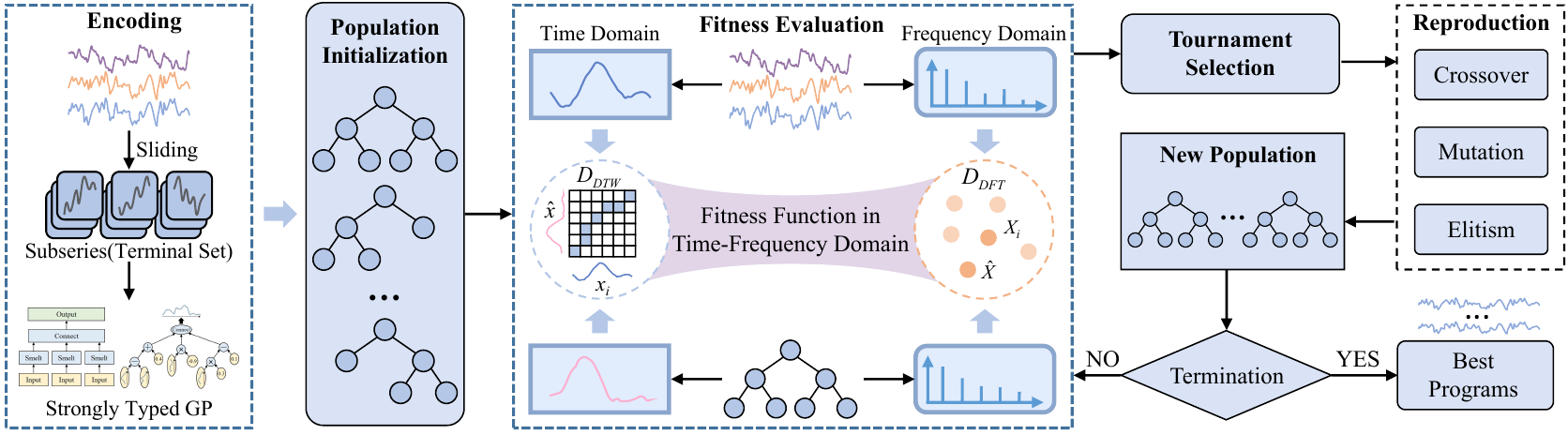}
  \caption{The framework of Evo-TFS. %The original dataset is oversampled using multi-population GP to balance the data.
  }

  \label{fig:framework}
\end{figure*}

The framework of Evo-TFS is illustrated in Fig.~\ref{fig:framework}. Evo-TFS involves multiple GP processes, each assigned to a specific target time-series sample that is selected from the original training set. Before an evolutionary learning process, a sliding window is used to segment all the time-series samples from the training set to extract possible subseries, which serve as the terminal set for GP. During each evolutionary process, the fitness function, which incorporates the time-frequency characteristics of that target sample, evaluates the quality of individuals. Once the termination criterion is satisfied, the evolutionary process stops, and a subset of high-quality individuals is selected as the final synthetic samples. These generated samples from all the GP processes are then combined with the original training dataset to become the final training set for developing classifiers.

\subsection{Tree Representation}

\begin{figure}
  \centering
  \includegraphics[width=0.5\textwidth]{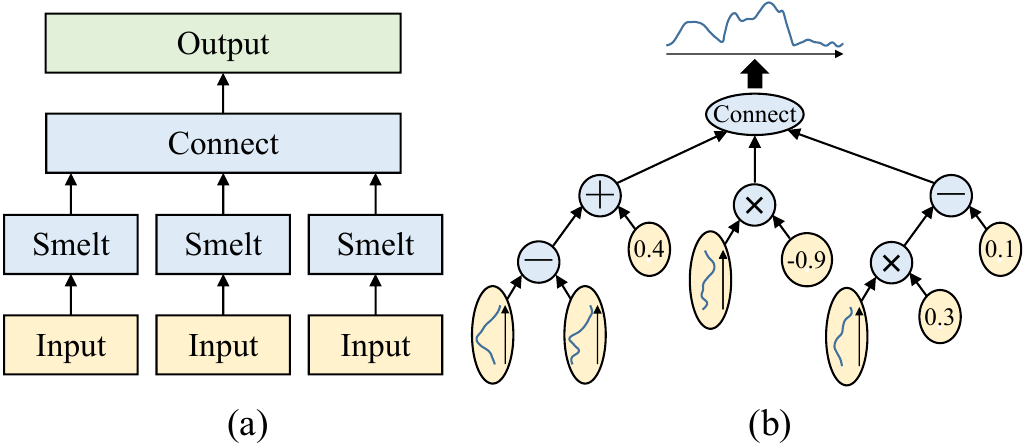}
  \caption{(a) The structure of STGP; (b) An example of an STGP program.}
  \label{fig:gp_representation}
\end{figure}

We have designed a new representation to generate time series data based on STGP. %To enhance the scalability of the algorithm, it uses STGP as the underlying framework. 
%STGP integrates functions and terminals with varying input and output types into a tree structure, ensuring type safety and reducing the likelihood of invalid program structures through enforced type constraints.
As shown in Fig.~\ref{fig:gp_representation}, an individual consists of four layers: $Input$, $Smelt$, $Connect$, and $Output$. The nodes in the $Input$ layer are selected from the terminal set, while the nodes in the $Smelt$ and $Connect$ layers are selected from the function set. The $Output$ layer produces the final sample and its corresponding label. The terminal and function sets are introduced as follows:

\begin{comment}
\begin{table}[htbp]
\caption{The Terminal Set and Function Set}
\label{tab:set}
\centering
\resizebox{\columnwidth}{!}{
\begin{tabular}{cccc}
\hline
 & Name & \multicolumn{2}{c}{Input Type} \\ \hline
\multirow{2}{*}{Terminal Set} & Subseries & \multicolumn{2}{c}{Array} \\
 & Random Constant & \multicolumn{2}{c}{Float} \\ \hline
 & Name & Input Type  & Output Type \\ \hline
\multirow{5}{*}{Function Set} & + & {[}Input,Input{]} & Input \\
 & - & {[}Input,Input{]} & Input \\
 & * & {[}Input,Input{]} & Input \\
 & / & {[}Input,Input{]} & Input \\
 & Connect & {[}Input(Array),...,Input(Array){]} & Output \\ \hline
\end{tabular}
}
\end{table}

\end{comment}

\begin{table}[htbp]
\caption{The Terminal Set and Function Set}
\label{tab:set}
\centering
\resizebox{\columnwidth}{!}{
\begin{tabular}{cccc}
\hline
 & Name & \multicolumn{2}{c}{Input Type} \\ \hline
\multirow{2}{*}{Terminal Set} & Subseries & \multicolumn{2}{c}{$Iput$ (Array)} \\
 & Random Constant & \multicolumn{2}{c}{$Iput$ (Float)} \\ \hline
 & Name & Input Type  & Output Type \\ \hline
\multirow{5}{*}{Function Set} & + & {[}$Iput$, $Iput${]} & $A\_Iput$ (Array)  \\
 & -- & {[}$Iput$, $Iput${]} & $A\_Iput$ (Array) \\
 & $\times$ & {[}$Iput$, $Iput${]} & $A\_Iput$ (Array)  \\
 & $\div$ & {[}$Iput$, $Iput${]} & $A\_Iput$ (Array)  \\
 & Connect & [$A\_Iput$, $A\_Iput$, $A\_Iput$]  & $Oput$ (Array) \\ \hline
\end{tabular}
}
\begin{flushleft}
\textbf{Note}: $Iput$, $A\_Iput$, and $Oput$ indicate different data types in an evolved tree program, and the data types shown in brackets after them represent their actual data types. 
\end{flushleft}
\end{table}

\subsubsection{\textbf{Terminal Set}}

%The terminal set is listed in TABLE~\ref{tab:set}, including subseries from all samples in the training set and a random constant uniformly distributed in the range from -1 to 1.

%The terminal set includes subseries from all the samples in the training set and a random constant uniformly distributed in the range from -1 to 1, reported in TABLE~\ref{tab:set}.

In each GP process, the terminal set includes all the possible subseries extracted using a sliding window from all the training time series data and a random constant uniformly distributed in the range from -1 to 1, reported in TABLE~\ref{tab:set}.

%Given a time series sample \( x_i = \{v_1, v_2, \dots, v_T\} \), where \( T \) represents the number of time steps in the sample, subseries \( S_i = \{s_{i,1}, s_{i,2}, \dots, s_{i,K}\} \) are extracted from each sample using a sliding window approach. Here, \( K \) denotes the number of subseries generated from each sample, and each subseries has a length of \( L \), i.e., \( s_{i,j} \in \mathbb{R}^L \). 

A time series sample is denoted as \( x_i = \{v_1, v_2, \dots, v_T\} \), where \( T \) represents the number of time steps in the sample. From each sample, the sliding window approach is used to extract subseries \( S_i = \{s_{i,1}, s_{i,2}, \dots, s_{i,K}\} \), where \( K \) denotes the number of subseries, and each subseries has a length of \( L \), i.e., \( s_{i,j} \in \mathbb{R}^L \). Specifically, for sample \( x_i \), the \( j \)-th subseries \( S_{i,j} = \{v_j, v_{j+1}, \dots, v_{j+L-1}\} \) is obtained by applying a sliding window of length \( L \), which starts at position \( j \).

For the entire training set, a subseries set can be represented as \( S = \{S_1, S_2, \dots, S_n\} \), where \( n \) is the number of training samples. Each set of subseries \( S_i \) contains \( K \) subseries, each with a length of \( L \). Therefore, the subseries set \( S \) has the dimensionality \( S \in \mathbb{R}^{N \times K \times L} \), where each training sample generates \( K \) subseries, and each subseries has a length of \( L \). To facilitate subsequent processing, the dimensionality of the subseries collection \( S \) is reshaped into \(S \in \mathbb{R}^{(N \times K) \times L} \), effectively flattening the subseries across all samples into a new collection containing \( N \times K \) subseries, each with a length of \( L \).

\subsubsection{\textbf{Function Set}}

TABLE~\ref{tab:set} reports the function set, including five functions, i.e., addition (+), subtraction (--), multiplication (\(\times\)), protected division ($\div$), and the $Connect$ function. Note that the protected division returns 1 when the denominator is 0. The four arithmetic functions (i.e., +, -, \(\times\), and protected $\div$) have two arguments, each of which is either a Float or an Array of length \( L \).
%and the input types for each argument are either Float or Array (of length \( L \)).
%The input types for these operators are either Float or Array (of length \( L \)). 4
%\textcolor{red}{When the two arguments has the same input types, the output type matches the input type. For example, if the inputs are \( [ \text{Float}, \text{Float} ] \), the output is of type Float. When the input types differ, the output type is determined by the higher-dimensional input type. For example, if the input is \( [ \text{Float}, \text{Array} ] \), the output is an Array type, with the left subtree output applied to each element of the array in the right subtree. }
When both arguments are arrays, an arithmetic function outputs an array based on the corresponding element-wise operation. %For example, if the inputs are \( [ \text{Float}, \text{Float} ] \), the output is of type Float. 
When one argument is of type array and the other is a float, an arithmetic function applies the operation between the float and each element of the array, returning the resulting array. %For example, if the input is \( [ \text{Float}, \text{Array} ] \), the output is an Array type, with the left subtree output applied to each element of the array in the right subtree. }
The $Connect$ function has three arguments, each of which is a subseries with the type of $A\_Iput$ from the $Smelt$ layer. This function returns a new series with the type of array.

Based on the terminal and function sets, the layers of a GP individual in Evo-TFS are introduced as follows:

\begin{itemize}
    \item The $Input$ layer: the nodes in this layer are selected from the terminal set and serve as the inputs to an individual.

     \item The $Smelt$ layer: this layer can select the functional nodes of +, --, \(\times\) and the protected ÷. This layer's functionality is to construct subseries.
     
     %The input types for these operators are either Float or Array (of length \( L \)). When the input types are the same, the output type matches the input type. For example, if the inputs are \( [ \text{Float}, \text{Float} ] \), the output is of type Float. \textcolor{red}{When the input types differ, the output type is determined by the higher-dimensional input type.} For example, if the input is \( [ \text{Float}, \text{Array} ] \), the output is an Array type, with the left subtree output applied to each element of the array in the right subtree. 

      \item The $Connect$ layer: the $Connect$ function is used in this layer. This layer's functionality is to concatenate the outputs from the $Smelt$ layer.  %Specifically, this function sequentially concatenates the outputs of the subtrees from left to right, aligning the dimensions of the output array with those of the original data.  As shown in Fig.~\ref{fig:gp_representation}(a) 

       \item The $Output$ layer: this layer produces the final sample and its corresponding label.
    
\end{itemize}

%The nodes in the $Input$ layer are drawn from the terminal set. The nodes in the $Smelt$ and $Connect$ layers are derived from the function set.  The $Output$ layer produces the final sample and its corresponding label.

\subsection{Individual Evaluation}

In GP, the fitness function is used to evaluate the quality of individuals in solving the given problem, thereby guiding the evolution towards better solutions. %The fitness function helps the algorithm assess the relative quality of different solutions, enabling appropriate choices during the optimization process. 
To comprehensively evaluate individual quality and ensure that it meets the desired requirements, a fitness function is designed to combine both local similarity and global characteristics of time series data, therefore taking into account information from both the time and frequency domains. 
\subsubsection{\textbf{Dynamic Time Warping (DTW) Distance}}

\begin{figure}
  \centering
  \includegraphics[width=0.3\textwidth]{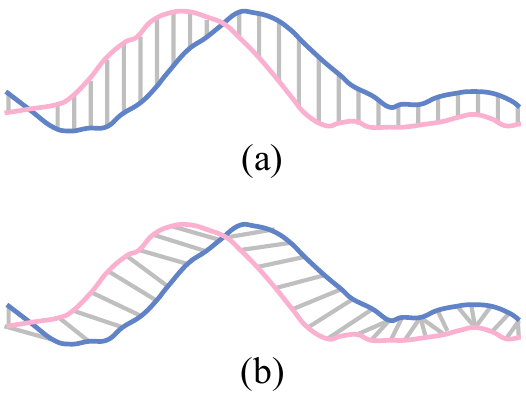}
  \caption{(a) Illustration of calculating the Euclidean distance between time series; (b) Illustration of calculating the DTW distance between time series.}
  \label{fig:dtw}
\end{figure}

DTW\cite{abanda2019review} is a method designed to measure the similarity between two time series, particularly in cases where nonlinear distortions exist along the time axis. As illustrated in Fig.~\ref{fig:dtw} (a), the traditional Euclidean distance assumes that two time series are strictly aligned along the time axis, meaning that similarity can only be evaluated when corresponding time points are matched in a one-to-one manner. %Consequently, Euclidean distance is highly sensitive to temporal misalignment and local distortions\cite{kate2016using}. Even slight time shifts between two sequences can lead to significant deviations in the computed similarity. 
Therefore, the Euclidean distance is highly sensitive to temporal misalignment and local distortions \cite{kate2016using}. As a result, even slight time shifts between two sequences can lead to significant deviations in the computed similarity, making Euclidean distance less robust and reliable for real-world time series comparisons.
As shown in Fig.~\ref{fig:dtw} (b), DTW accommodates nonlinear temporal distortions by employing a dynamic programming approach to identifying the optimal alignment path that minimizes the overall accumulated distance. %\textcolor{orange}{This method allows for flexible point-wise alignment, effectively mitigating the effects of local temporal misalignments and improving the \textcolor{red}{robustness} of similarity measurements. Since DTW determines the optimal alignment across the entire time series, it excels in handling temporal shifts, local variations, and periodic distortions.}

To compute the DTW distance between two time series samples $x = \{v_1, v_2, \dots, v_m\}$ and $\hat{x} = \{\hat{v}_1, \hat{v_2}, \dots, \hat{v}_n\}$, we first construct a cost matrix $ D \in \mathbb{R}^{m \times n} $, where each element is defined as:

\begin{equation}
D(i, j) = \| v_i - \hat{v}_j \|.
\label{eq_3}
\end{equation}
Next, we define the accumulated cost matrix $C \in \mathbb{R}^{m \times n}$, where the first element is initialized as $C(1,1) = D(1,1)$. For \( i \geq 2 \) and \( j \geq 2 \), the recurrence relation for \( C(i,j) \) is given by:

\begin{equation}
C(i, j) = D(i, j) + \min \begin{cases} 
C(i-1, j), \\ 
C(i, j-1), \\ 
C(i-1, j-1). 
\end{cases}
\label{eq_4}
\end{equation}
Finally, the DTW distance between \( x \) and \( \hat{x}\) is defined as:
\begin{equation}
D_{\text{DTW}}(x,\hat{x}) = C(m, n).
\label{eq_5}
\end{equation}
%In this study, the generated sample $\hat{x}$ is strictly aligned in length with the target sample $x_i$, ensuring that \( m = n \).
In this study, the generated sample $\hat{x}$ and the target sample $x$ are strictly aligned in length, ensuring that \( m = n \).

%\textcolor{orange}{DTW is capable of capturing subtle temporal differences between generated and target samples.} %Through minimizing the individual's DTW distance, Evo-TFS progressively guides the generated samples to approximate the target sample. 
Using DTW in Evo-TFS allows for flexible point-wise alignment to capture subtle temporal differences between generated and target samples.
Compared with Euclidean distance or other similarity measures, the DTW distance is more robust to local distortions and temporal shifts in time series data. This capability enables DTW to effectively align sequences with temporal misalignments and provide a more accurate measure of similarity between samples. In particular, DTW can identify change patterns across different time steps and capture local-level similarities, making it well-suited to handle the complex distributions of real-world time series data.
\subsubsection{\textbf{Fourier Distance}}

\begin{figure}
  \centering
  \includegraphics[width=0.45\textwidth]{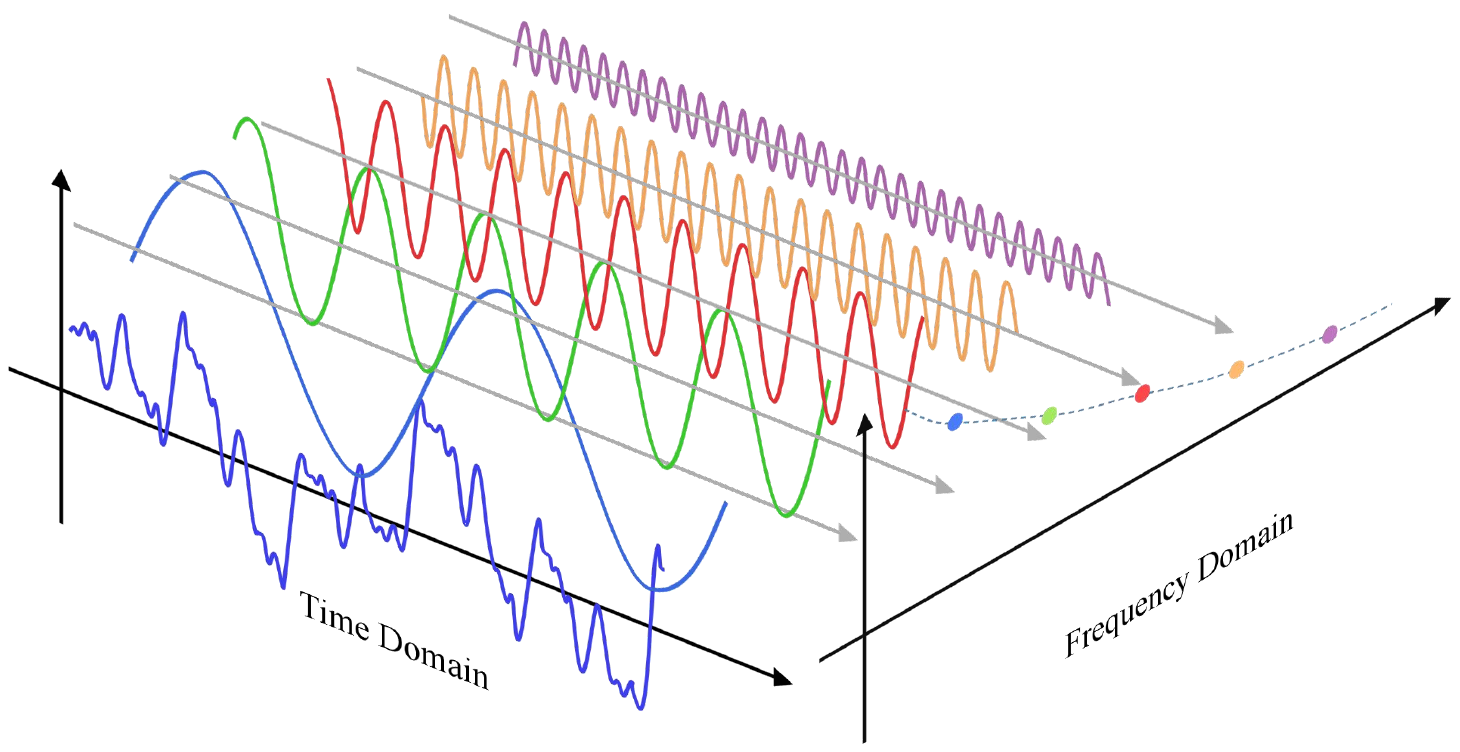}
  \caption{Illustration of Discrete Fourier Transform.}
  \label{fig:fft}
\end{figure}

Discrete Fourier Transform (DFT) \cite{sundararajan2024discrete} is an important mathematical tool used to analyze the frequency components of a finite-length discrete time series. By transforming a time-domain sample into its frequency-domain representation using DFT, the sample's constituent frequency components can be effectively identified. Specifically, as shown in Fig.~\ref{fig:fft}, DFT decomposes a sample into a set of sinusoidal or cosine components, each with a specific frequency, amplitude, and phase. %This process enables DFT to reveal the frequency characteristics of the sample. 

Given a discrete time series $x$ of length $T$, DFT transforms the sample into its frequency-domain representation $X[k]$, where \( k = 0, 1, 2, \dots, T-1 \) represents different frequency components. The formula for DFT is as follows \cite{sundararajan2024discrete}:
\begin{equation}
X[k] = \sum_{n=0}^{T-1} x(n) e^{-j \frac{2\pi}{T} k n}, \quad k = 0, 1, \dots, T-1
\label{eq_6}
\end{equation}

where:
\begin{itemize}
    \item \( x(n) \) is the \( n \)-th time point of the original time-domain sample,
    \item \( X[k] \) is the complex representation of the sample's \( k \)-th frequency component in the frequency domain,
    \item \( e^{-j \frac{2\pi}{T} k n} \) is the complex exponential function that represents the frequency response of the sine and cosine components.
\end{itemize}

%Fourier distance is commonly used to compare the similarity between two time series in the frequency domain, with the total Fourier distance, which combines both magnitude and phase information, being the most commonly used metric.  The total Fourier distance is typically calculated by considering the differences between the magnitude and phase spectra. The formula for calculating the distance between two time series, $x_i$ and $\hat{x}$, is as follows:

The Fourier distance is commonly used to compare the similarity between two time series in the frequency domain. %The total Fourier distance combines both magnitude and phase information, becoming the most commonly used metric. 
%The total Fourier distance is typically calculated by considering the differences between the magnitude and phase spectra. 
The formula for calculating the distance between two time series, $x$ and $\hat{x}$, is as follows \cite{1705539}:

\begin{equation}
\scriptsize
D_{DFT}(x, \hat{x}) = \sqrt{
\sum_{k=0}^{T-1} ( |X[k]| - |\hat{X}[k]| )^2 + 
\sum_{k=0}^{T-1} ( \arg(X[k]) - \arg(\hat{X}[k]) )^2,
}
\label{eq_7}
\end{equation}
where:  
\begin{itemize}
    \item $X[k]$ and $\hat{X}[k]$ are frequency domain values of $x$ and $\hat{x}$, respectively;
    \item $|X[k]|$ and $|\hat{X}[k]|$ are the magnitudes of the samples $x$ and $\hat{x}$ at the frequency point $k$;  
    \item $\arg(X[k])$ and $\arg(\hat{X}[k])$ are the phases of the samples $x$ and $\hat{x}$ at the frequency point $k$. 
\end{itemize}

Using the Fourier distance in Evo-TFS captures the spectral differences between a generated sample and the target sample. %guiding the generated sample towards the target sample's spectral features by minimizing the Fourier distance during the evolutionary process. 
The total Fourier distance measures the frequency-domain similarity between samples by considering both magnitude and phase spectrum differences, especially when the time series exhibits periodic or quasi-periodic variations. %Specifically, using Fourier distance could capture the global spectral features .%, providing a powerful tool for measuring similarity from a frequency-domain perspective.

\subsubsection{\textbf{The Fitness Function}}
Based on the DTW and Fourier distance metrics, the fitness function is defined as follows:

%\begin{equation}
%Fitness = \alpha \cdot Q(D_{\text{DTW}}(x_i,\hat{x})) + (1 - \alpha) \cdot Q(D_{\text{DFT}}(x_i,\hat{x})),
%\label{eq_1}
%\end{equation}

\begin{equation}
\textcolor{black}{Fitness =  \alpha Q(D_{\text{DTW}}(x,\hat{x})) +  (1-\alpha) Q(D_{\text{DFT}}(x,\hat{x})),}
\label{eq_1}
\end{equation}

\begin{equation}
Q(x) = e^{\frac{-x^2}{2\sigma^2}},
\label{eq_2}
\end{equation}
where $x$ represents the target sample, $\hat{x}$ represents a generated sample, and \textcolor{black}{$\alpha \in [0, 1]$ is a weighting parameter used to balance the importance of the time-domain distance and frequency-domain distance.} $D_{DTW}$ denotes the DTW distance between the generated sample and the target sample, while $D_{DFT}$ represents the total Fourier distance between the generated sample and the target sample. Finally, $Q(x)$ represents a partial function of a Gaussian function to normalize $D_{DTW}$ and $D_{DFT}$ to make them in the same range. %The range of $\alpha$ is [0,1]. Since the information in the time domain is as important as that in the frequency domain, we set $\alpha$ = 0.5 in this article.

The fitness function integrates the DTW and Fourier distance metrics to more comprehensively assess the similarity between a generated sample and the target sample. DTW distance effectively captures local variations within time windows, particularly in cases with time axis distortions. This offers a flexible matching method that ensures that the generated time series sample aligns with the target sample at the local feature level. On the other hand, the Fourier distance operates in the frequency domain, revealing variations in the frequency components of the time series data and helping capture global structural features. By combining these two distance metrics, the fitness function is able to simultaneously capture local fluctuations and preserve the overall patterns in the time series, thus enhancing the diversity, plausibility, and representativeness of the generated sample.

%During each evolutionary process, the fitness function, which incorporates the time-frequency characteristics of that target sample, evaluates the quality of individuals. Once the termination criterion is satisfied, the evolutionary process stops, and a subset of high-quality individuals is selected as the final synthetic samples. These generated samples from all the GP processes are then combined with the original training dataset to become the final training set for developing classifiers. 

\textcolor{black}{
\subsection{Evolutionary Process}
Evo-TFS employs multiple independent GP processes, each using a single target time-series sample from the minority class and evolving a population of candidate solutions over a number of generations. This enables Evo-TFS to generate diverse samples by allowing diverse existing time-series data from the minority class to be used for generating new samples. During each evolutionary process, the fitness function assesses the quality of individuals, determining how closely their time-frequency characteristics align with those of the target time-series sample.}

\textcolor{black}{
The number of GP processes in Evo-TFS is defined as follows:
\begin{equation}
N_p = \sum_{i \in C}\min(N_{gi},N_i),
\label{eq_8}
\end{equation}
where $C$ represents the set of labels, $N_i$ represents the number of samples in Class $i$, and $N_{gi}$ represents the number of samples to be generated for the minority class $i$. }

\textcolor{black}{
The imbalance ratio ($IR$) can be used to calculate how many instances are required to be generated for the minority class \cite{10793073}. 
Let $N_{pi}$ denote the number of processes used to generate samples for class $i$. 
\begin{itemize}
    \item If $IR = 2$, then $N_{pi} = N_{gj}$. In this case, all existing samples are used as target samples, each assigned to a separate GP process that returns the best individual as a generated sample.
    \item If \( IR > 2 \), all existing samples are used as the target samples, each assigned to a separate GP process that needs to return the top \( \frac{N_{gi}}{N_{pi}} \) optimal solutions as the generated samples.  
    \item If $IR < 2$, then \( N_{pi} < N_i \). In this case, the \( N_{pi} \) samples closest to the class center (introduced later in more detail) are selected as the target samples, each assigned to a separate GP process that ultimately returns the best individual as a generated sample.  
\end{itemize}
}

\textcolor{black}{The class center of a class is defined as the element-wise mean of all time series belonging to the same class. For class $i$, the class center (denoted as $X_c$) is computed as follows:}

\begin{equation}
\textcolor{black}{X_c(t) = \frac{1}{N_i} \sum_{j=1}^{N_i} x_j(t), \quad \forall t = 1, \dots, T}
\label{eq_c}
\end{equation}
\textcolor{black}{where $x_j(t)$ denotes the $t$-th time point of $x_j$, and $T$ denotes the number of time steps in a sample. }

\textcolor{black}{The detailed steps of calculating the class center are introduced as follows. First, we applied Min-Max normalization to the time-series data to unify the scale of all samples.
Afterwards, since random fluctuations and high-frequency noise tend to cancel out during the element-wise summation, the resulting mean vector serves as a stable and smooth prototype. %This allows it to effectively capture the ``average behavior" of the class while minimizing the impact of idiosyncratic noise found in individual samples.
}

\textcolor{black}{Note that the proposed method is not limited to periodic time series, as it does not rely on any prior assumption of periodicity or cyclic structure. The segmentation strategy adopted in our proposed method simply divides each time-series data into sub-series using a fixed-length sliding window, without attempting to align with any specific period. Note that for non-periodic data, dividing into sub-series would not affect the quality of the generated samples. This is because each independent GP process maintains a population of candidate solutions (i.e., generated samples), each of which is evaluated according to the fitness function to indicate its quality, i.e., whether a generated sample effectively matches the target time-series sample in terms of time-frequency characteristics.  A generated instance with a low fitness value, indicating poor quality, is unlikely to survive during the evolutionary process.}

\textcolor{black}{This paper focuses specifically on the class imbalance problem in time-series classification. When applying the proposed method to datasets with missing values, appropriate imputation techniques must be used as a preprocessing step. The proposed method is inherently robust when applied to imbalanced datasets containing anomalies for two reasons as follows. First, the proposed method explicitly prioritizes samples closest to the class centroid when generating synthetic instances. Second, GP has the built-in ability to automatically determine which subseries from a sample in the training set will be used to generate a new sample. If a synthetic GP-represented sample is constructed from a subseries of outliers or noisy data, it will likely receive a low fitness score. As a result, its probability of being selected for survival is greatly reduced, which inherently mitigates the negative impacts of outliers or noisy data on the overall quality of the generated data. }

\section{Experimental Design}

To validate the effectiveness of Evo-TFS, we carried out extensive experiments on ten time series datasets and conducted a thorough comparative analysis with \textcolor{black}{six} representative sampling methods. %Through systematic comparative analysis with five state-of-the-art baseline methods, the experimental results demonstrate the superior performance of our proposed approach.
\subsection{Datasets}

In our experiments, we used \textcolor{black}{12} time series datasets from the UCR Archive\footnote{These datasets are available at: \href{https://www.cs.ucr.edu/~eamonn/time_series_data/}{\url{https://www.cs.ucr.edu/~eamonn/time_series_data/}}}, with detailed information provided in TABLE~\ref{tab:datasets} (\textcolor{black}{DistalPhalanxOutlineAgeGroup is abbreviated as DistalP-AgeGrp, and MiddlePhalanxOutlineCorrect is abbreviated as MiddleP-Corr}). These datasets cover a wide range of domains, including healthcare, geoscience, industrial inspection, biology, energy management, and motion recognition. Since all the datasets have been split into training and test sets, we retained the original splits.
Following \cite{zhao2022t}, we performed stratified sampling on the training sets of the selected datasets to produce new datasets with diverse $IR$. The resulting $IR$ range from 2.00 to \textcolor{black}{38.80}, covering a wide spectrum of imbalance levels. In the sampled datasets, the majority class is labeled as {\bf{Maj}} and the minority class as {\bf{Min}}.

\begin{table}[h]
\caption{\textcolor{black}{Datasets in the experiments
}}
\label{tab:datasets}
\centering
\begin{tabular}{cccccc}
\hline
Dataset & No.sample & Class & IR & Length & Index \\ \hline
ShapeletSim & 15 & 2 & 2.00 & 500 & D1 \\
ECG200 & 100 & 2 & 2.23 & 96 & D2 \\
Ham & 76 & 2 & 3.00 & 431 & D3 \\
Herring & 52 & 2 & 3.00 & 512 & D4 \\
WormsTwoClass & 135 & 2 & 3.50 & 270 & D5 \\
ItalyPowerDemand & 41 & 2 & 4.86 & 24 & D6 \\
\color{black}DistalP-AgeGrp & 317 & 3 & 8.57 & 80 & D7 \\
PowerCons & 99 & 2 & 10.00 & 144 & D8 \\
ScreenType & 149 & 3 & 10.42 & 720 & D9 \\
Yoga & 147 & 2 & 13.70 & 426 & D10 \\
Computers & 131 & 2 & 20.83 & 720 & D11 \\
\color{black}MiddleP-Corr & 398 & 2 & 38.80 & 80 & D12 \\ \hline
\end{tabular}
\end{table}

\subsection{Baseline Methods}
%We compare our method with 4 oversampling methods for static data and 1 oversampling method for time series data. 

The baseline methods are described as follows:
\begin{itemize}
    \item \textbf{SMOTE \cite{chawla2002smote}:} It is the most commonly-used oversampling method that synthesizes new samples by interpolating between a {\bf{Min}} sample and its neighbors to balance data.
    \item \textbf{ADASYN \cite{he2008adasyn}:} Different from SMOTE, it prioritizes the difficult-to-learn samples in {\bf{Min}} to generate more synthetic samples to balance data.   
    %based on the density of data distributions, where the {\bf{Min}} instances located in low-density regions generate more synthetic instances to balance data.
    \item \textbf{Borderline-SMOTE1 (abbreviated as B1SMOTE) \cite{han2005borderline}}: It synthesizes new samples by interpolating borderline samples (labeled as $danger$ samples) from {\bf{Min}} with their same-class (minority) neighbors.
    \item \textbf{Borderline-SMOTE2 (abbreviated as B2SMOTE) \cite{han2005borderline}}: Same as Borderline-SMOTE1, it synthesizes new samples based on borderline samples in {\bf{Min}} and their same-class neighbors. However, it further considers neighbors from {\bf{Maj}} in some cases when interpolating, generating synthetic samples that are often closer to {\bf{Min}} but located within the decision boundary region.
    \item \textbf{T-SMOTE \cite{zhao2022t}}: It incorporates time information during the generation process of synthetic samples to ensure that the generated data retains its time dependence. It also assigns different weights to each {\bf{Min}} sample according to its importance to generate new samples. 
    \item \textcolor{black}{\textbf{Diffusion-TS\cite{yuan2024diffusionts}}: It is a novel diffusion-based method that generates multivariate time series samples by using an encoder-decoder transformer with disentangled temporal representations. Guided by a decomposition technique, it captures the overarching semantics of the data, while the Transformer extracts detailed sequential patterns from the noisy input.}

\end{itemize}

\subsection{Parameter Settings}

TABLE~\ref{tab:parameter} presents the parameter configurations for the proposed Evo-TFS method. Evo-TFS generates $N_g$ synthetic samples through $N_p$ GP processes, each of which generates $Max(1, \frac{N_g}{N_p})$ samples. 
To balance the trade-off between model performance and computational efficiency, we set the population size for a certain dataset based on $IR$. If $IR < 15$, the population size in each GP process is set to 30; otherwise, it is set to 50. The number of generations is set to 50 to ensure a sufficient number of evaluations. In each GP process, individuals are initialized using the ramped half-and-half method. Regarding the configuration of genetic operators, the crossover rate and mutation rate are set to 80\% and 20\%\cite{10120936}, respectively. The tournament selection is used, with a tournament size of 3. To prevent tree bloating during the learning process, the maximum tree depth is set to 10. For the Gaussian function $Q(x)$ involved in Evo-TFS, the standard deviation parameter $\sigma$ is set to 10.

\begin{table}[]
\caption{PARAMETER SETTINGS}
\label{tab:parameter}
\centering
\normalsize
\begin{tabular}{c c}
\hline
Parameter                          & Value                \\ \hline 
%The number of GP processes           & $N_p$\\
%The number of samples$/$Population  & Max(1,\( \frac{N_g}{N_p} \))   \\
%Population size in a GP process         & 30$/$50       \\
The number of generations        & 50 \\     
Tournament size                    & 3                  \\  
Rate of crossover  & 80\%                \\ 
Rate of mutation                           & 20\%                 \\ 
The number of elites & 2\\
Initialization & Ramped-half-and-half \\
Maximal tree depth        & 10                   \\
\textcolor{black}{$\alpha$} & \textcolor{black}{0.5}\\
%$\sigma$ in the fitness function     & 10    \\ 
\hline
\end{tabular}
\end{table}

\begin{table*}

\caption{\textcolor{black}{The results of the sampling methods with CNN in the time domain on the 12 UCR datasets. The best results are in \textbf{bold}, and the second best are {\ul underlined}}}
\centering

\label{tab:cnn_result}
\setlength{\tabcolsep}{3pt}
\renewcommand\arraystretch{1.6}

\begin{minipage}{1\textwidth}  % Adjust the width of the table
  \raggedright  % Use \raggedleft if you want to align the table to the right
  \resizebox{\textwidth}{!}{%
\begin{tabular}{|c|c|cccccccccccc|}
\hline
Metric & Methods & D1 & D2 & D3 & D4 & D5 & D6 & D7 & D8 & D9 & D10 & D11 & D12 \\ \hline
  & SMOTE & 0.110±0.181 (+)& {\ul 0.852±0.019 (=)}& 0.704±0.015 (+)& 0.461±0.077 (+)& 0.680±0.020 (=)& 0.878±0.007 (+)& 0.642±0.023 (+)& 0.852±0.046 (+)& 0.252±0.022 (+)& {\ul 0.355±0.023 (-)}& 0.106±0.028 (+)& 0.767±0.005 (+)\\
  & ADASYN & 0.048±0.106 (+)& 0.838±0.018 (+)& 0.711±0.019 (+)& 0.420±0.061 (+)& 0.694±0.017 (=)& 0.875±0.007 (+)& 0.641±0.023 (+)& 0.774±0.042 (+)& 0.251±0.031 (+)& 0.259±0.018 (-)& 0.099±0.025 (+)& 0.777±0.007 (=)\\
  & B1SMOTE & 0.125±0.228 (+)& 0.834±0.013 (+)& 0.698±0.029 (+)& 0.448±0.085 (+)& 0.695±0.024 (=)& 0.876±0.007 (+)& 0.649±0.030 (+)& 0.805±0.047 (+)& 0.311±0.028 (+)& 0.191±0.029 (=)& 0.138±0.036 (=)& \textbf{0.787±0.009 (-)}\\
  & B2SMOTE & 0.048±0.1006 (+)& 0.821±0.015 (+)& 0.710±0.023 (+)& 0.454±0.081 (+)& 0.687±0.017 (=)& 0.875±0.007 (+)& 0.650±0.025 (+)& 0.765±0.040 (+)& 0.241±0.017 (+)& \textbf{0.360±0.026 (-)}& 0.122±0.026 (+)& 0.773±0.009 (+)\\
  & T-SMOTE & 0.249±0.185 (+)& 0.843±0.014 (+)& {\ul 0.777±0.027 (-)}& 0.312±0.056 (+)& {\ul 0.697±0.035 (=)}& 0.867±0.0104 (+)& 0.588±0.042 (+)& \textbf{0.925±0.026 (=)}& 0.317±0.023 (+)& 0.255±0.027 (-)& 0.117±0.036 (+)& 0.727±0.001 (+)\\
 & Diffusion-TS & {\ul 0.261 (+)} & 0.839 (+) & \textbf{0.803 (-)} & {\ul 0.495 (+)} & \textbf{0.727 (-)} & {\ul 0.907 (=)} & {\ul 0.668 (+)} & 0.883 (=) & {\ul 0.349 (=)} & 0.229 (=) & {\ul 0.145 (=)} & 0.727 (+) \\
 & Evo-TFS & \textbf{0.390±0.220} & \textbf{0.857±0.024} & 0.753±0.041 & \textbf{0.563±0.074} & 0.693±0.025 & \textbf{0.910±0.018} & \textbf{0.699±0.042} & {\ul 0.902±0.051} & \textbf{0.351±0.028} & 0.208±0.063 & \textbf{0.158±0.054} & {\ul 0.780±0.008} \\ \cline{2-14} 

\multirow{-8}{*}{F1-Score} & Total & \multicolumn{12}{c|}{49 +, 15 =, 8 -} \\ \hline
  & SMOTE & 0.136±0.160 (+) & 0.804±0.021 (=) & 0.649±0.031 (+) & 0.551±0.060 (+) & 0.496±0.036 (=)& 0.885±0.006 (+) & 0.769±0.016 (=) & 0.862±0.041 (+) & 0.299±0.034 (+) & {\ul 0.461±0.017 (-)} & 0.234±0.036 (+) & 0.508±0.033 (+) \\
 & ADASYN & 0.089±0.130 (+) & 0.805±0.022 (=) & 0.652±0.029 (+) & 0.519±0.048 (+) & 0.524±0.033 (=)& 0.881±0.006 (+) & 0.770±0.013 (=) & 0.795±0.036 (+) & 0.299±0.049 (+) & 0.382±0.015 (-) & 0.226±0.031 (+) & 0.523±0.030 (+) \\
 & B1SMOTE & 0.079±0.150 (+) & {\ul 0.808±0.018 (=)} & 0.653±0.040 (+) & 0.538±0.065 (+) & {\ul 0.529±0.036 (-)} & 0.883±0.007 (+) & 0.774±0.017 (=) & 0.821±0.040 (+) & 0.387±0.037 (+) & 0.323±0.027 (=)& {\ul 0.270±0.039 (=)}& \textbf{0.568±0.036 (=)} \\
 & B2SMOTE & 0.094±0.137 (+) & 0.794±0.027 (+)& 0.651±0.026 (+) & 0.545±0.063 (+) & 0.526±0.032 (=)& 0.882±0.006 (+) & 0.773±0.016 (=) & 0.788±0.034 (+) & 0.281±0.027 (+) & \textbf{0.465±0.020 (-)} & 0.254±0.030 (+) & 0.530±0.037 (=) \\
 & T-SMOTE & \textbf{0.323±0.170 (=)} & 0.801±0.017 (=) & 0.734±0.040 (=) & 0.436±0.048 (+) & \textbf{0.535±0.034 (=)}& 0.875±0.009 (+) & 0.721±0.028 (+) & 0.897±0.024 (=)& 0.400±0.031 (+) & 0.380±0.022 (-)& 0.246±0.041 (=)& 0.510±0.015 (+) \\
 & Diffusion-TS & 0.295 (=) & 0.795 (+) & \textbf{0.755 (=)} & {\ul 0.581 (+)} & 0.528 (=) & {\ul 0.911 (=)} & 0.768 (=) & {\ul 0.904 (=)} & {\ul 0.403 (+)} & 0.342 (=) & 0.265 (=) & 0.545 (=) \\
 & Evo-TFS & {\ul 0.315±0.183} & \textbf{0.816±0.028} & {\ul 0.745±0.047} & \textbf{0.626±0.061} & 0.503±0.045 & \textbf{0.914±0.017} & \textbf{0.775±0.030} & \textbf{0.907±0.046} & \textbf{0.443±0.041} & 0.336±0.055 & \textbf{0.288±0.058} & {\ul 0.562±0.085} \\ \cline{2-14}
\multirow{-8}{*}{G-Mean} & Total & \multicolumn{12}{c|}{38 +, 29 =, 5 -} \\ \hline
\end{tabular}
}
\end{minipage}
\end{table*}

Evo-TFS was implemented using the DEAP package \cite{fortin2012deap}, and the baseline sampling methods were implemented using the imbalanced-learn package \cite{lemaavztre2017imbalanced}. To ensure comparison fairness in the experiments, this study adopts a unified assessment framework, which is introduced as follows. First, a classifier is trained using a rebalanced training set generated by both Evo-TFS and the baseline methods, followed by performance assessment on the corresponding test set.

Specifically, we employ two classifiers to evaluate and analyze the time-series data: 
\begin{itemize}    
\item[(1)]  \textcolor{black}{Due to the superior performance of CNN in TSC tasks, we use a CNN to learn the time-domain features~\cite{zhao2017convolutional}. Note that the CNN model we used in our study is a 1D convolutional neural network (1D-CNN)~\cite{10.1007/978-3-319-08010-9_33}, which is specifically designed for time-series data processing. The architecture of 1D-CNN utilizes 1D convolution kernels to slide over the time axis, effectively capturing local temporal patterns and trends.    } 
\item[ (2)] To further investigate the impact of data sampling on classification in the frequency domain, we introduce a method that combines the Fourier Transform and Random Forest (RF) to learn the frequency-domain features~\cite{rigatti2017random}.
\end{itemize}

Finally, a metric is proposed to evaluate the density consistency among different classes of a rebalanced dataset, defined as follows:
% Finally, a metric has been normally employed to evaluate the uniformity of the rebalanced dataset, defined as follows:

\begin{equation}
% \scriptsize
U = |\frac{\displaystyle \sum_{i \in Maj}\sum_{j \in n_{i}^{k}} D_{DTW}(i,j)}{k \times |Maj|} - \frac{\displaystyle\sum_{i \in Min}\sum_{j \in n_{i}^{k}} D_{DTW}(i,j)}{k \times |Min|}|,
\label{eq_9}
\end{equation}
where $n_i^k$ denotes the $k$ nearest neighbors of sample $i$. The value of $U$ ranges from $0$ to $+\infty$, where a smaller value indicates a more uniform distribution between the majority and minority classes after re-balancing.

\begin{table*}
\caption{\textcolor{black}{The results of the sampling methods with RF in the frequency domain on the 12 UCR datasets. The best results are in \textbf{bold}, and the second best are {\ul underlined}}}
\centering

\label{tab:dft_rf_result}
\setlength{\tabcolsep}{3pt}
\renewcommand\arraystretch{1.6}

\begin{minipage}{1\textwidth}  % Adjust the width of the table
  \raggedright  % Use \raggedleft if you want to align the table to the right
  \resizebox{\textwidth}{!}{%
\begin{tabular}{|c|c|cccccccccccc|}
\hline
Metric & Methods & D1 & D2 & D3 & D4 & D5 & D6 & D7 & D8 & D9 & D10 & D11 & D12 \\ \hline
  & SMOTE & 0.492±0.057 (+) & 0.808±0.010 (+) & 0.212±0.059 (+)& 0.547±0.014 (=)& 0.728±0.052 (-) & \textbf{0.749±0.020 (-)}& 0.641±0.013 (=) & 0.509±0.009 (=)& {\ul 0.174±0.005 (-)} & {\ul 0.300±0.001 (-)}& 0.348±3E-04 (+) & 0.734±0.003 (+) \\
 & ADASYN & 0.492±0.057 (+) & 0.798±0.011 (+) & 0.212±0.059 (+)& \textbf{0.550±0.015 (=)}& 0.729±0.050 (-) & 0.706±0.004 (-)& \textbf{0.654±0.013 (=)} & 0.435±0.004 (+) & {\ul 0.174±0.006 (-)} & {\ul 0.300±6E-04 (-)}& 0.348±5E-04 (+) & 0.737±0.002 (+) \\
 & B1SMOTE & 0.514±0.061 (+) & 0.787±0.009 (+) & 0.168±0.057 (+) & 0.546±0.013 (=)& 0.651±0.066 (=)& 0.707±0.006 (-)& 0.645±0.017 (=) & 0.509±0.004 (=)& 0.166±1E-04 (=)& 0.292±7E-04 (+) & 0.347±0.001 (+) & \textbf{0.740±0.002 (=)} \\
 & B2SMOTE & 0.512±0.066 (+) & 0.845±0.011 (-)& 0.235±0.054 (+)& \textbf{0.550±0.014 (=)}& \textbf{0.745±0.032 (-)} & 0.705±0.004 (-)& 0.642±0.013 (=) & 0.418±0.004 (+) & \textbf{0.177±0.008 (-)} & 0.290±0.001 (+) & 0.347±3E-04 (+) & 0.731±0.004 (+)\\
 & T-SMOTE & 0.452±0.086 (+) & {\ul 0.846±0.008 (-)}& {\ul 0.224±0.028 (+)}& 0.545±2E-06 (=)& {\ul 0.731±0.036 (-)} & {\ul 0.746±0.017 (-)}& 0.547±0.019 (+) & 0.490±0.016 (+) & 0.166±0.002 (=)& 0.291±3E-04 (+) & 0.345±4E-06 (+) & 0.726±6E-06 (+) \\
 & Diffusion-TS & {\ul 0.555 (+)} & \textbf{0.884 (-)} & 0.127 (+) & 0.525 (=) & 0.409 (+) & 0.658 (-) & 0.641 (+) & {\ul 0.531 (=)} & 0.166 (=) & \textbf{0.302 (-)} & \textbf{0.352 (=)} & 0.726 (+) \\
 & Evo-TFS & \textbf{0.595±0.052} & 0.822±0.017 & \textbf{0.421±0.013}  & 0.533±0.052 & 0.677±0.038 & 0.637±0.045 & {\ul 0.653±0.019} & \textbf{0.533±0.060} & 0.169±0.006 & 0.298±3E-4 & {\ul 0.351±0.005} & \textbf{0.740±0.8E-4} \\ \cline{2-14} 
\multirow{-8}{*}{F1-Score} & Total & \multicolumn{12}{c|}{34 +, 19 =, 19 -} \\ \hline
& SMOTE & 0.568±0.043 (=)& 0.803±0.019 (+)& {\ul 0.308±0.054 (=)} & 0.306±0.084 (=)& 0.664±0.042 (-) & 0.585±0.071 (=)& \textbf{0.769±0.011 (-)} & 0.424±0.039 (=) & 0.025±0.047 (=) & \textbf{0.076±0.015 (=)} & 0.086±0.016 (+)& 0.333±0.020 (+)\\
 & ADASYN & 0.568±0.043 (=)& 0.782±0.015 (+) & 0.238±0.054 (+) & 0.305±0.080 (+)& 0.585±0.041 (=)& 0.420±0.022 (+) & {\ul 0.767±0.011 (-)} & 0.280±0.028 (+) & \textbf{0.037±0.051 (-)}& 0.071±0.010 (=) & 0.080±0.027 (+)& 0.274±0.019 (+)\\
 & B1SMOTE & 0.585±0.046 (=)& 0.775±0.015 (+) & 0.298±0.056 (+) & 0.256±0.101 (+)& 0.601±0.053 (=)& 0.423±0.026 (+) & 0.763±0.013 (-) & 0.234±0.03 (+) & 0.004±0.011 (=) & 0.051±0.016 (+) & 0.090±0.034 (+)& 0.283±0.021 (+)\\
 & B2SMOTE & 0.584±0.050 (=)& 0.801±0.021 (+)& 0.270±0.046 (+) & 0.339±0.063 (=)& \textbf{0.678±0.027 (-)} & 0.416±0.021 (+) & 0.764±0.010 (-) & 0.305±0.029 (+) & {\ul 0.022±0.044 (=)} & 0.053±0.010 (+) & 0.091±0.007 (+)& 0.333±0.014 (+)\\
 & T-SMOTE & 0.539±0.067 (+)& 0.807±0.015 (+)& 0.262±0.024 (+) & 0.325±0.016 (=)& {\ul 0.666±0.029 (-)} & 0.574±0.052 (=)& 0.683±0.014 (-) & 0.418±0.038 (=) & 0.008±0.020 (=) & 0.066±0.001 (+) & 0.001±5E-04 (+)& {\ul 0.348±0.021 (=)}\\
 & Diffusion-TS & \textbf{0.616 (=)} & \textbf{0.832 (-)} & 0.258 (+) & {\ul 0.369 (=)} & 0.510 (+) & {\ul 0.591 (=)} & 0.659 (=) & {\ul 0.430 (=)} & 0.000 (=) & 0.061 (+) & {\ul 0.126 (=)} & 0.333 (+) \\
 & Evo-TFS & {\ul 0.590±0.055} & {\ul 0.819±0.022} & \textbf{0.330±0.021}  & \textbf{0.374±0.096} & 0.589±0.053 & \textbf{0.594±0.055} & 0.663±0.017 & \textbf{0.456±0.084} & 0.003±0.013 & {\ul 0.075±0.001} & \textbf{0.131±0.019} & \textbf{0.351±0.021} \\ \cline{2-14} 

\multirow{-8}{*}{G-Mean} & Total & \multicolumn{12}{c|}{34 +, 28 =, 10 -} \\ \hline
\end{tabular}
}
\end{minipage}
\end{table*}

\section{Results and Analysis}

On each dataset, Evo-TFS and baseline methods (except Diffusion-TS) were independently executed 30 times using different random seeds. Two classifiers were trained on the training sets rebalanced by different sampling methods, and tested on the same test set. \textcolor{black}{The performance was evaluated using F1-Score~\cite{liu2019model, yin2020novel}, G-Mean~\cite{he2009learning} and AUC~\cite{fawcett2006introduction} measures (the detailed AUC results are provided in \textbf{Appendix D}). The Wilcoxon signed-rank test with a significance level of 0.05 and the Holm–Bonferroni correction~\cite{lee2018proper} for multiple comparisons have been conducted to assess whether the performance difference between Evo-TFS and a baseline method is statistically significant \cite{demvsar2006statistical}.}
\subsection{Overall Results}

\textcolor{black}{TABLES~\ref{tab:cnn_result} and ~\ref{tab:dft_rf_result} report the F1-Score and G-Mean of the two classifiers trained based on Evo-TFS and the six baseline methods on the test sets. In these tables, ``+'' indicates significantly better performance of Evo-TFS compared with a baseline method, ``--'' indicates significantly worse performance, and ``='' indicates no significant difference relative to a baseline. In the row of ``Evo-TFS", the values after `$\pm$'' denote the corresponding standard deviations.
}

%As shown in TABLE~\ref{tab:cnn_result}, when using the CNN classifier in the time domain, Evo-TFS significantly outperforms the five baseline sampling methods in terms of both F1-Score and G-Mean. Specifically, Evo-TFS (Best) achieves the highest average F1-Score of 0.679, surpassing all the baseline methods, followed by Evo-TFS (Mean) with an average score of 0.579, which is also notably superior to the competing methods. On a per-dataset basis, Evo-TFS (Best) obtains the highest F1-Score across all 10 datasets, while Evo-TFS (Mean) secures second-best results on six datasets. Regarding G-Mean, Evo-TFS (Best) again attains the highest average score of 0.687 and ranks first on all datasets. Evo-TFS (Mean) follows closely with an average G-Mean of 0.589, outperforming all baselines and achieving six second-best results. These findings highlight the strong generalization capability of Evo-TFS in handling class imbalance.

\begin{table}

\caption{\textcolor{black}{The statistical results of significance tests on F1-Score and G-Mean}}
\renewcommand\arraystretch{1}
\label{tab:method_comparison}
\centering
\begin{tabular}{@{}lcccc@{}}
\toprule
\textbf{Metric} & \multicolumn{2}{c}{\textbf{F1-Score}} & \multicolumn{2}{c}{\textbf{G-Mean}} \\
\cmidrule(lr){2-3} \cmidrule(lr){4-5}
\textbf{Methods} & \textbf{CNN (+/=/--)} & \textbf{RF (+/=/--)} & \textbf{CNN (+/=/--)} & \textbf{RF (+/=/--)} \\
\midrule
\textbf{SMOTE} & 9/2/1 & 5/3/4 & 8/3/1 & 3/7/2 \\
\textbf{ADASYN} & 9/2/1 & 6/2/4 & 8/3/1 & 7/3/2 \\
\textbf{B1SMOTE} & 8/3/1 & 5/6/1 & 6/5/1 & 8/3/1 \\
\textbf{B2SMOTE} & 10/1/1 & 6/2/4 & 8/3/1 & 7/3/2 \\
\textbf{T-SMOTE} & 8/2/2 & 7/2/3 & 5/6/1 & 5/5/2 \\
\textbf{Diffusion-TS} & 5/5/2 & 5/4/3 & 3/9/0 & 4/7/1 \\ \hline
\textbf{Total} & 49/15/8 & 34/19/19 & 38/29/5 & 34/28/10 \\ 
\bottomrule
\end{tabular}
\end{table}

\textcolor{black}{As shown in TABLE~\ref{tab:cnn_result}, when using the CNN classifier in the time domain, Evo-TFS outperforms six baseline sampling methods in terms of both F1-Score and G-Mean. Specifically, based on the average results over the 30 runs, Evo-TFS achieves the best performance on 7 datasets for each metric, ranking either first or second in all the 12 datasets except for three in F1-Score and two in G-Mean.} 
\textcolor{black}{Based on the statistical significance tests, Evo-TFS achieves significantly better F1-Score performance in 49 out of the total 72 cases, while performing significantly worse in only 8. Similarly, in terms of the G-Mean measure, Evo-TFS achieves significantly better results in 39 out of the 72 cases.} These results validate the effectiveness of Evo-TFS in generating time-series data to address the class imbalance issue for CNN in TSC tasks.

\textcolor{black}{As shown in TABLE~\ref{tab:dft_rf_result}, Evo-TFS also demonstrates a clear advantage when used with the RF classifier in the frequency domain. For the F1-Score, Evo-TFS achieves the best or second-best performance on 6 out of 12 datasets, one more than Diffusion-TS (i.e., 5). 
For G-Mean, Evo-TFS outperforms the baseline methods on 6 datasets and ranks second on 3 datasets. Statistical analysis further confirms these observations. In terms of F1-Score, Evo-TFS is significantly better or statistically similar in 53 out of the 72 cases. For G-Mean, Evo-TFS performs significantly better or statistically similar in 62 cases, and is significantly worse in only 10 cases. These results demonstrate that Evo-TFS is effective in rebalancing time-series data, contributing to a significant improvement in the performance of traditional classifiers in the frequency domain.}

In summary, Evo-TFS demonstrates significant advantages in generating time-series data for TSC, with both the CNN classifier in the time domain and the RF classifier in the frequency domain. %Evo-TFS not only achieves leading results in key metrics such as F1-Score and G-Mean but also maintains stable performance across diverse datasets. The statistical significance analysis further supports the reliability of these outcomes, underscoring the generalizability and effectiveness of Evo-TFS as a sampling strategy across various learning settings.

\subsection{Summary on the Results and Analysis}

\begin{table*}[]
\caption{\textcolor{black}{The performance ranking of Evo-TFS (mean) and the baseline method (mean) on two classifiers on 12 datasets. The best rankings are in \textbf{bold}, and the second best are {\ul underlined}}}
\centering
\label{tab:rank_mean}
\setlength{\tabcolsep}{3pt}
\renewcommand\arraystretch{1.3}

\begin{minipage}{1\textwidth}  % Adjust the width of the table
  \raggedright  % Use \raggedleft if you want to align the table to the right
  \resizebox{\textwidth}{!}{%
\begin{tabular}{|cc|ccccccccccccc|ccccccccccccc|}
\hline
\multicolumn{2}{|c|}{Metric} & \multicolumn{13}{c|}{F1-Score} & \multicolumn{13}{c|}{G-Mean} \\ \hline
\multicolumn{1}{|c|}{Classifier} & Methods & D1 & D2 & D3 & D4 & D5 & D6 & D7 & D8 & D9 & D10 & D11 & D12 & Average & D1 & D2 & D3 & D4 & D5 & D6 & D7 & D8 & D9 & D10 & D11 & D12 & Average \\ \hline
\multicolumn{1}{|c|}{\multirow{7}{*}{CNN}} & SMOTE & 5.0 & 2.0 & 6.0 & 3.0 & 7.0 & 3.0 & 5.0 & 4.0 & 5.0 & 2.0 & 6.0 & 5.0 & 4.42 & 4.0 & 4.0 & 7.0 & 3.0 & 7.0 & 3.0 & 5.0 & 4.0 & 5.0 & 2.0 & 6.0 & 7.0 & 4.75 \\
\multicolumn{1}{|c|}{} & ADASYN & 7.0 & 5.0 & 4.0 & 6.0 & 4.0 & 6.0 & 6.0 & 6.0 & 6.0 & 3.0 & 7.0 & 3.0 & 5.25 & 6.0 & 3.0 & 5.0 & 6.0 & 5.0 & 6.0 & 4.0 & 6.0 & 6.0 & 3.0 & 7.0 & 5.0 & 5.17 \\
\multicolumn{1}{|c|}{} & B1SMOTE & 4.0 & 6.0 & 7.0 & 5.0 & 3.0 & 4.0 & 4.0 & 5.0 & 4.0 & 7.0 & 3.0 & 1.0 & 4.42 & 7.0 & 2.0 & 4.0 & 5.0 & 2.0 & 4.0 & 2.0 & 5.0 & 4.0 & 7.0 & 2.0 & 1.0 & 3.75 \\
\multicolumn{1}{|c|}{} & B2SMOTE & 6.0 & 7.0 & 5.0 & 4.0 & 6.0 & 5.0 & 3.0 & 7.0 & 7.0 & 1.0 & 4.0 & 4.0 & 4.92 & 5.0 & 7.0 & 6.0 & 4.0 & 4.0 & 5.0 & 3.0 & 7.0 & 7.0 & 1.0 & 4.0 & 4.0 & 4.75 \\
\multicolumn{1}{|c|}{} & T-SMOTE & 3.0 & 3.0 & 2.0 & 7.0 & 2.0 & 7.0 & 7.0 & 1.0 & 3.0 & 4.0 & 5.0 & 6.0 & 4.17 & 1.0 & 5.0 & 3.0 & 7.0 & 1.0 & 7.0 & 7.0 & 3.0 & 3.0 & 4.0 & 5.0 & 6.0 & 4.33 \\
\multicolumn{1}{|c|}{} & Diffusion-TS & 2.0 & 4.0 & 1.0 & 2.0 & 1.0 & 2.0 & 2.0 & 3.0 & 2.0 & 5.0 & 2.0 & 7.0 & {\ul 2.75} & 3.0 & 6.0 & 1.0 & 2.0 & 3.0 & 2.0 & 6.0 & 2.0 & 2.0 & 5.0 & 3.0 & 3.0 & {\ul 3.17} \\
\multicolumn{1}{|c|}{} & Evo-TFS & 1.0 & 1.0 & 3.0 & 1.0 & 5.0 & 1.0 & 1.0 & 2.0 & 1.0 & 6.0 & 1.0 & 2.0 & \textbf{2.08} & 2.0 & 1.0 & 2.0 & 1.0 & 6.0 & 1.0 & 1.0 & 1.0 & 1.0 & 6.0 & 1.0 & 2.0 & \textbf{2.08} \\ \hline
\multicolumn{1}{|c|}{\multirow{7}{*}{RF}} & SMOTE & 5.5 & 5.0 & 4.5 & 3.0 & 4.0 & 3.0 & 6.0 & 4.0 & 3.0 & 3.0 & 3.0 & 4.0 & 4.00 & 5.5 & 4.0 & 2.0 & 5.0 & 3.0 & 3.0 & 5.0 & 3.0 & 6.0 & 1.0 & 5.0 & 5.0 & 3.96 \\
\multicolumn{1}{|c|}{} & ADASYN & 5.5 & 6.0 & 4.5 & 1.0 & 3.0 & 6.0 & 1.0 & 6.0 & 2.0 & 2.0 & 4.0 & 3.0 & {\ul 3.67} & 5.5 & 6.0 & 7.0 & 6.0 & 6.0 & 6.0 & 6.0 & 6.0 & 1.0 & 3.0 & 6.0 & 7.0 & 5.46 \\
\multicolumn{1}{|c|}{} & B1SMOTE & 3.0 & 7.0 & 6.0 & 4.0 & 6.0 & 5.0 & 3.0 & 3.0 & 6.0 & 5.0 & 5.0 & 2.0 & 4.58 & 3.0 & 7.0 & 3.0 & 7.0 & 4.0 & 5.0 & 3.0 & 7.0 & 4.0 & 7.0 & 4.0 & 6.0 & 5.00 \\
\multicolumn{1}{|c|}{} & B2SMOTE & 4.0 & 3.0 & 2.0 & 2.0 & 1.0 & 7.0 & 4.0 & 7.0 & 1.0 & 7.0 & 6.0 & 5.0 & 4.08 & 4.0 & 5.0 & 4.0 & 3.0 & 1.0 & 7.0 & 4.0 & 5.0 & 2.0 & 6.0 & 3.0 & 4.0 & 4.00 \\
\multicolumn{1}{|c|}{} & T-SMOTE & 7.0 & 2.0 & 3.0 & 5.0 & 2.0 & 4.0 & 7.0 & 5.0 & 5.0 & 6.0 & 7.0 & 7.0 & 5.00 & 7.0 & 3.0 & 5.0 & 4.0 & 2.0 & 4.0 & 1.0 & 4.0 & 3.0 & 4.0 & 7.0 & 2.0 & 3.83 \\
\multicolumn{1}{|c|}{} & Diffusion-TS & 2.0 & 1.0 & 7.0 & 7.0 & 7.0 & 1.0 & 5.0 & 2.0 & 7.0 & 1.0 & 1.0 & 6.0 & 3.92 & 1.0 & 1.0 & 6.0 & 2.0 & 7.0 & 2.0 & 7.0 & 2.0 & 7.0 & 5.0 & 2.0 & 3.0 & {\ul 3.75} \\
\multicolumn{1}{|c|}{} & Evo-TFS & 1.0 & 4.0 & 1.0 & 6.0 & 5.0 & 2.0 & 2.0 & 1.0 & 4.0 & 4.0 & 2.0 & 1.0 & \textbf{2.75} & 2.0 & 2.0 & 1.0 & 1.0 & 5.0 & 1.0 & 2.0 & 1.0 & 5.0 & 2.0 & 1.0 & 1.0 & \textbf{2.00} \\ \hline
\end{tabular}
}
\end{minipage}
\end{table*}

TABLE~\ref{tab:method_comparison} summarizes the statistical significance test results. % comparing the proposed Evo-TFS method with five baseline methods under the F1-Score and G-Mean metrics.
Overall, Evo-TFS demonstrates a clear advantage in most cases.
\textcolor{black}{Compared with the baseline sampling methods, Evo-TFS helped CNN achieve significantly better or statistically similar performance in 63 cases (F1-Score) and 66 cases (G-Mean) out of the total 72.
    Similarly, when applied to RF, Evo-TFS assisted in achieving significantly better or statistically similar performance in 53 cases (F1-Score) and 62 cases (G-Mean) out of the total 72.}
\textcolor{black}{From a method-specific perspective, Evo-TFS demonstrates stronger performance when applied with CNN compared to RF, especially in comparison to baseline methods such as SMOTE, and ADASYN, etc. In contrast, Evo-TFS with RF generally achieves better performance than Diffusion-TS.} In summary, in terms of both F1-Score and G-Mean metrics, Evo-TFS achieves significantly better performance in most cases. These results highlight the robustness and stability of the proposed method across different classifiers and evaluation metrics.

\begin{table}[]
\caption{\textcolor{black}{The results of the ablation study on the 4 datasets}}
\label{tab:ablation}
\centering
\resizebox{\columnwidth}{!}{
\begin{tabular}{|c|c|c|ccc|}
\hline
Datasets & Metric & Methods & w/o DTW & w/o DFT & Evo-TFS \\ \hline
\multirow{4}{*}{ECG200} & \multirow{2}{*}{F1-Score} & CNN & {\ul 0.855} & 0.854 & \textbf{0.857} \\
 &  & RF & \textbf{0.847} & 0.802 & {\ul 0.822} \\ \cline{2-6} 
 & \multirow{2}{*}{G-Mean} & CNN & 0.806 & \textbf{0.819} & {\ul 0.816} \\
 &  & RF & \textbf{0.827} & 0.815 & {\ul 0.819} \\ \hline
\multirow{4}{*}{ItalyPowerDemand} & \multirow{2}{*}{F1-Score} & CNN & \textbf{0.913} & 0.901 & {\ul 0.910} \\
 &  & RF & 0.630 & {\ul 0.632} & \textbf{0.637} \\ \cline{2-6} 
 & \multirow{2}{*}{G-Mean} & CNN & 0.896 & \textbf{0.924} & {\ul 0.914} \\
 &  & RF & 0.586 & {\ul 0.589} & \textbf{0.594} \\ \hline
\multirow{4}{*}{PowerCons} & \multirow{2}{*}{F1-Score} & CNN & 0.893 & {\ul 0.898} & \textbf{0.902} \\
 &  & RF & {\ul 0.528} & 0.525 & \textbf{0.533} \\ \cline{2-6} 
 & \multirow{2}{*}{G-Mean} & CNN & \textbf{0.908} & 0.897 & {\ul 0.907} \\
 &  & RF & \textbf{0.460} & 0.419 & {\ul 0.456} \\ \hline
\multirow{4}{*}{Computers} & \multirow{2}{*}{F1-Score} & CNN & \textbf{0.167} & \textbf{0.167} & 0.158 \\
 &  & RF & {\ul 0.333} & {\ul 0.333} & \textbf{0.351} \\ \cline{2-6} 
 & \multirow{2}{*}{G-Mean} & CNN & {\ul 0.281} & 0.275 & \textbf{0.288} \\
 &  & RF & {\ul 0.129} & 0.127 & \textbf{0.131} \\ \hline
\end{tabular}
}
\end{table}

\begin{table}[]

\caption{\textcolor{black}{The results of the  parameter sensitivity on the 4 datasets}}
\label{tab:sensitivity}
\centering
\resizebox{\columnwidth}{!}{
\begin{tabular}{|c|c|c|lll|}
\hline
Datasets & Metric & Methods & $\alpha=0.3$ & $\alpha=0.7$ & $\alpha=0.5$ \\ \hline
\multirow{4}{*}{ECG200} & \multirow{2}{*}{F1-Score} & CNN & 0.851 & {\ul 0.854} & \textbf{0.857} \\
 &  & RF & 0.819 & \textbf{0.828} & {\ul 0.822} \\ \cline{2-6} 
 & \multirow{2}{*}{G-Mean} & CNN & 0.812 & \textbf{0.819} & {\ul 0.816} \\
 &  & RF & \textbf{0.824} & {\ul 0.823} & 0.819 \\ \hline
\multirow{4}{*}{ItalyPowerDemand} & \multirow{2}{*}{F1-Score} & CNN & \textbf{0.912} & 0.903 & {\ul 0.910} \\
 &  & RF & 0.633 & \textbf{0.639} & {\ul 0.637} \\ \cline{2-6} 
 & \multirow{2}{*}{G-Mean} & CNN & \textbf{0.919} & 0.907 & {\ul 0.914} \\
 &  & RF & \textbf{0.602} & {\ul 0.599} & 0.594 \\ \hline
\multirow{4}{*}{PowerCons} & \multirow{2}{*}{F1-Score} & CNN & 0.897 & {\ul 0.899} & \textbf{0.902} \\
 &  & RF & 0.527 & {\ul 0.530} & \textbf{0.533} \\ \cline{2-6} 
 & \multirow{2}{*}{G-Mean} & CNN & \textbf{0.911} & {\ul 0.909} & 0.907 \\
 &  & RF & 0.453 & \textbf{0.459} & {\ul 0.456} \\ \hline
\multirow{4}{*}{Computers} & \multirow{2}{*}{F1-Score} & CNN &  \textbf{0.158} & 0.157 & \textbf{0.158} \\
 &  & RF & {\ul 0.333} & {\ul 0.333} & \textbf{0.351} \\ \cline{2-6} 
 & \multirow{2}{*}{G-Mean} & CNN & {\ul 0.283} & 0.281 & \textbf{0.288} \\
 &  & RF & \textbf{0.133} & 0.130 & {\ul 0.131} \\ \hline
\end{tabular}
}
\end{table}

\textcolor{black}{As shown in TABLE~\ref{tab:method_comparison}, Evo-TFS outperforms the baseline methods on most imbalanced time series datasets. This is mainly because GP is capable of generating synthetic time series that capture intricate temporal patterns by learning complex transformations from existing data. Moreover, both time-domain and frequency-domain information are incorporated into the fitness function, guiding the evolution toward better solutions that maintain the temporal and spectral integrity of the original data. 
However, in a few cases, Evo-TFS performs worse than the baseline methods. This typically arises when minority-class instances are relatively concentrated in the feature space. In this case, both SMOTE (and its variants) with its local interpolation and Diffusion-TS with its distribution learning are able to generate compact, representative samples, leading to competitive performance.}

To investigate whether Evo-TFS can generate high-quality balanced datasets to improve the classifier's performance, TABLE~\ref{tab:rank_mean} summarize the detailed rankings of each method across the 12 datasets for each classification algorithm. 
% TABLE~\ref{tab:rank_best} shows the ranking comparison between Evo-TFS (Best) and the baseline methods across the 10 datasets, two classifiers, and two performance metrics. Among the 40 ranking cases (10 datasets $\times$ 2 classifiers $\times$ 2 metrics), Evo-TFS achieves \textbf{38 first-place rankings} and \textbf{2 second-place rankings}. 
% TABLE~\ref{tab:rank_mean} summarizes the average ranking results of Evo-TFS and the baseline methods. 
\textcolor{black}{In total, Evo-TFS obtains 24 first-place and 12 second-place rankings out of the 48 cases. Moreover, it achieves the best average ranking across both the two classifiers and two evaluation metrics.} These results demonstrate that the proposed Evo-TFS method is capable of generating high-quality balanced datasets, effectively boosting the performance of various classifiers across diverse domains in imbalanced TSC tasks.  %Regarding the average ranks, Evo-TFS achieves scores of 2.40 and 2.60 on F1-Score with CNN and RF, respectively, and 2.40 and 2.30 on G-Mean, surpassing all the baseline methods. 
%Regarding the average ranks, Evo-TFS attains score of 2.40 and 2.60 on F1-Score with CNN and RF, respectively, and 2.40 and 2.30 on G-Mean, consistently surpassing all the baseline methods. 
%These results indicate that Evo-TFS delivers consistently strong performance across domains and classifiers, demonstrating its robustness in handling class imbalance.

\begin{figure*}
  \centering
  \includegraphics[width=1.0\textwidth]{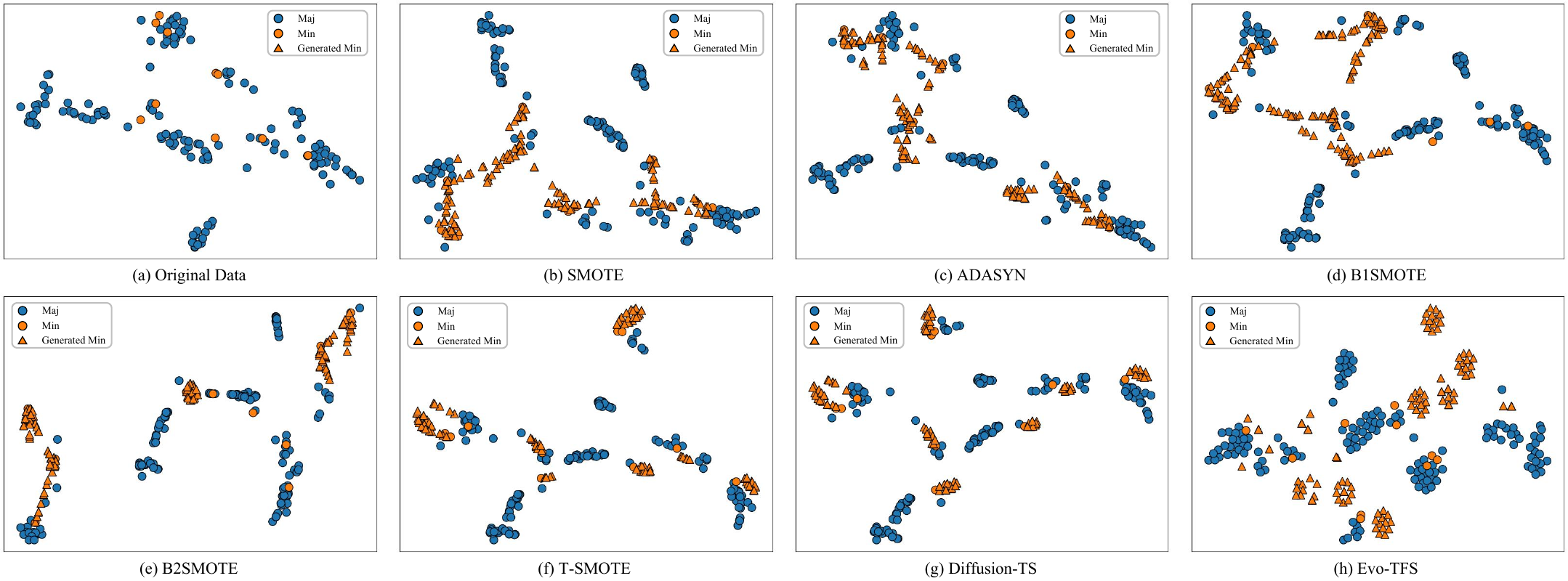}
  \caption{\textcolor{black}{The visual scatter plot reduced to 2D using the t-SNE method on the Screentype dataset.}}
  \label{fig:scatter_screentype}
\end{figure*}

\begin{figure*}
  \centering
  \includegraphics[width=1.0\textwidth]{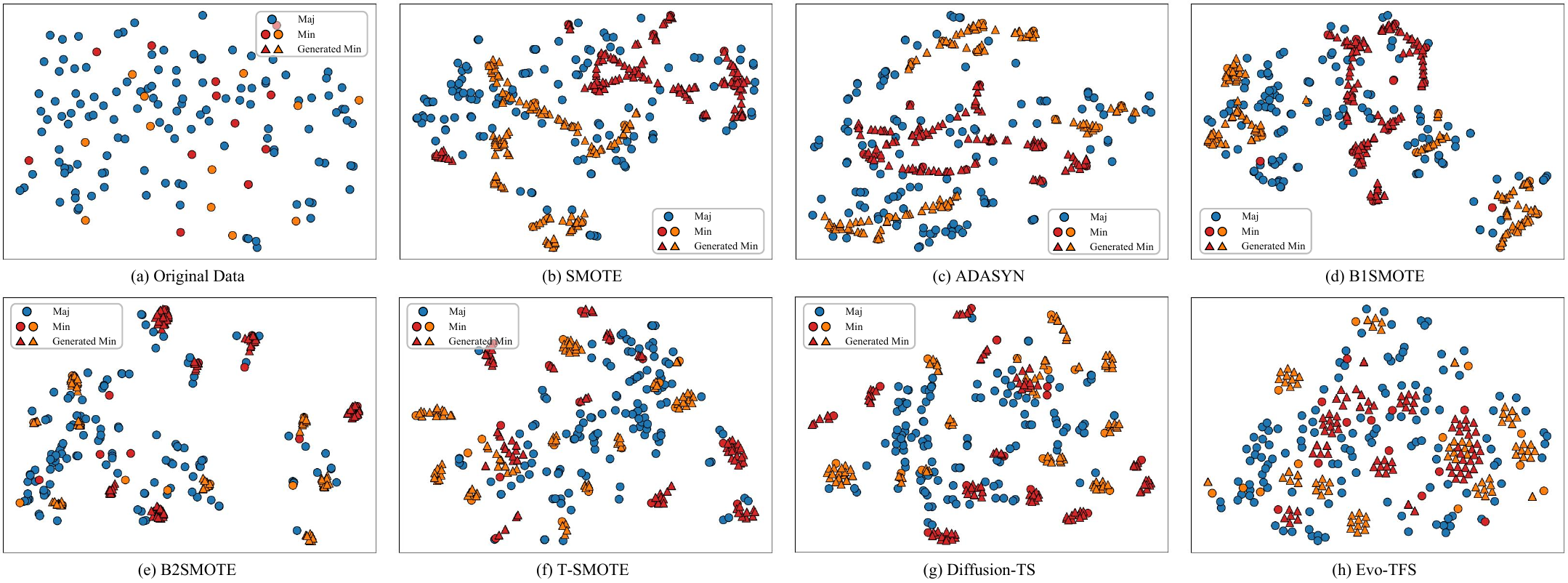}
  \caption{\textcolor{black}{The visual scatter plot reduced to 2D using the t-SNE method on the Yoga dataset.}}
  \label{fig:scatter_Yoga}
\end{figure*}

\subsection{\textcolor{black}{Ablation Studies and Parameter Sensitivity Analysis}}

\textcolor{black}{We further examine whether using both the DTW and DFT distances is necessary for improving performance. In the ablation study, removing DTW from the fitness function is denoted as w/o DTW, and removing DFT is denoted as w/o DFT.}

\textcolor{black}{As shown in TABLE \ref{tab:ablation}, removing either DTW or DFT is detrimental to performance stability across different datasets and classifiers. This is because DTW and DFT capture complementary information. DTW emphasizes local temporal alignment, while DFT captures global frequency patterns. 
The inclusion of both components in the fitness function enables Evo-TFS to achieve results that are more robust across different metrics and datasets.}

\textcolor{black}{To further examine how the parameter $\alpha$ in the fitness function influences the results, TABLE~\ref{tab:sensitivity} summarizes the outcomes under different $\alpha$ settings. As shown in TABLE \ref{tab:sensitivity}, the performance differences across different values of $\alpha$ are very small on all datasets and for both classifiers. These observations indicate that Evo-TFS is not sensitive to the parameter $\alpha$.
}

\subsection{Data Visualization and Further Analysis}

\textcolor{black}{
Figs.~\ref{fig:scatter_screentype} and ~\ref{fig:scatter_Yoga} present t-SNE visualizations of oversampled data distributions on the Screentype and Yoga datasets, respectively. As indicated in these figures, interpolation-based sampling methods, e.g., SMOTE and ADASYN, tend to produce samples with linear distributions, usually struggling to model complex temporal dynamics. This is mainly because these methods operate solely in the feature space, ignoring the sequential order and dependencies of time series data. By generating synthetic samples through linear interpolation between randomly selected neighbors, it is very likely to corrupt the inherent temporal structure, such as trends and autocorrelation, thereby generating noisy data points. T-SMOTE and Diffusion-TS (Figs.~\ref{fig:scatter_screentype} and ~\ref{fig:scatter_Yoga}(f-g)) produce concentrated cluster-like data blocks that better preserve local data structures, yet exhibiting insufficient global coverage capability that hinders boundary information. As shown in Figs.~\ref{fig:scatter_screentype} and ~\ref{fig:scatter_Yoga}(h), the synthetic samples generated by Evo-TFS exhibit more well-structured and evenly distributed characteristics. These samples not only maintain intra-class consistency but also effectively expand the boundary regions of minority classes, filling sparse areas with continuity and structural consistency. %This highlights the ability of Evo-TFS to better model and transfer structural features from time-series data. 
}

\begin{figure*}
  \centering
  \includegraphics[width=0.95\textwidth]{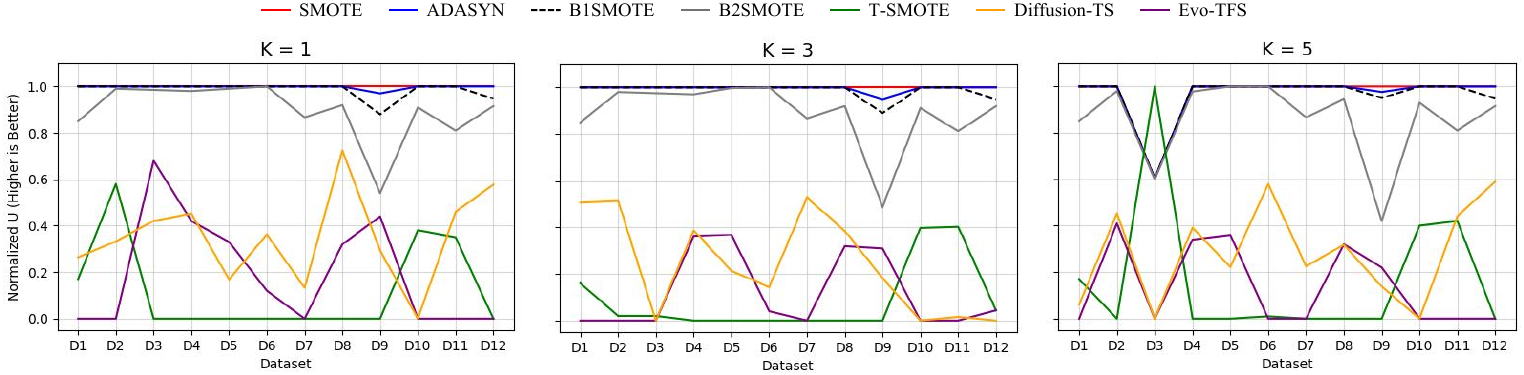}
  \caption{\textcolor{black}{The normalized density consistency indicators of different methods under different $K$ values.}}
  \label{fig:Uniformity_Index}
\end{figure*}

% Fig. \ref{fig:Uniformity_Index} provides additional quantitative evidence comparing the distribution uniformity between synthesized minority-class samples and original majority-class samples under
Fig. \ref{fig:Uniformity_Index} provides further quantitative evidence for the density consistency of different classes in a rebalanced dataset by a sampling method.
These density consistency scores ($U$) were normalized to $[0, 1]$ for each dataset, with smaller values reflecting better performance. %Ideally, a well-balanced dataset should exhibit similar \textcolor{red}{density} between the majority and minority classes, i.e., $U$ approaching 0. 
Ideally, a well-balanced dataset is expected to exhibit similar density between the majority and minority classes, i.e., $U$ approaching 0. 
The results indicate that traditional interpolation-based sampling methods, such as SMOTE and ADASYN, frequently generate minority-class samples with densities that significantly differ from those of the majority class. \textcolor{black}{Differently, samples generated by T-SMOTE, Diffusion-TS, and Evo-TFS tend to exhibit density distributions that are more similar to the majority class.} Notably, Evo-TFS achieves the highest density consistency in most cases, confirming its advantage in generating high-quality, representative samples for imbalanced TSC tasks.

In summary, by integrating spatial distribution insights from both t-SNE visualizations and uniformity statistics, Evo-TFS demonstrates clear superiority in preserving data structure, improving class balance, and supporting robust downstream classification performance.

\section{Conclusions}

%This study aims to generate samples that preserve temporal characteristics and exhibit diversity, in order to balance class distributions in time series data and improve classifier performance. 

This study aims to generate a diverse set of time-series samples that retain temporal characteristics, intending to balance data distributions and enhance classifier performance. To achieve this, we proposed a time-frequency domain-based evolutionary oversampling method. Specifically, we designed a GP framework for oversampling, where STGP was used to constrain input and output types. Moreover, time-frequency domain information was integrated into the fitness function to effectively guide the evolutionary process.

Experiments on the imbalanced time series datasets demonstrated that the proposed Evo-TFS method outperforms the baseline methods by generating time-series samples that are higher in quality, more diverse, and more evenly distributed. Besides, Evo-TFS outperforms these baseline methods on both time-domain and frequency-domain classifiers. While Evo-TFS demonstrates strong capability in generating high-quality, diverse samples, the use of GP results in high computational costs and long training times. In the future, we will explore strategies to accelerate the process.

\ifCLASSOPTIONcaptionsoff
  \newpage
\fi

\bibliographystyle{IEEEtran}
\bibliography{reference}

% Generated by IEEEtran.bst, version: 1.14 (2015/08/26)
\begin{thebibliography}{10}
\providecommand{\url}[1]{#1}
\csname url@samestyle\endcsname
\providecommand{\newblock}{\relax}
\providecommand{\bibinfo}[2]{#2}
\providecommand{\BIBentrySTDinterwordspacing}{\spaceskip=0pt\relax}
\providecommand{\BIBentryALTinterwordstretchfactor}{4}
\providecommand{\BIBentryALTinterwordspacing}{\spaceskip=\fontdimen2\font plus
\BIBentryALTinterwordstretchfactor\fontdimen3\font minus \fontdimen4\font\relax}
\providecommand{\BIBforeignlanguage}[2]{{%
\expandafter\ifx\csname l@#1\endcsname\relax
\typeout{** WARNING: IEEEtran.bst: No hyphenation pattern has been}%
\typeout{** loaded for the language `#1'. Using the pattern for}%
\typeout{** the default language instead.}%
\else
\language=\csname l@#1\endcsname
\fi
#2}}
\providecommand{\BIBdecl}{\relax}
\BIBdecl

\bibitem{LI2025104025}
\BIBentryALTinterwordspacing
X.~Li, W.~Li, X.~Yu, Z.~Han, and Q.~Jin, ``Financial risk assessment of imbalanced data based on nonlinear causal time-series network,'' \emph{Information Processing \& Management}, vol.~62, no.~3, p. 104025, 2025. [Online]. Available: \url{https://www.sciencedirect.com/science/article/pii/S0306457324003844}
\BIBentrySTDinterwordspacing

\bibitem{2020Utilizing}
Y.~Wang, Y.~Wei, H.~Yang, J.~Li, and Q.~Wu, ``Utilizing imbalanced electronic health records to predict acute kidney injury by ensemble learning and time series model,'' \emph{BMC Medical Informatics and Decision Making}, vol.~20, no.~1, p. 238, 2020.

\bibitem{jiang2019gan}
W.~Jiang, Y.~Hong, B.~Zhou, X.~He, and C.~Cheng, ``A {GAN}-based anomaly detection approach for imbalanced industrial time series,'' \emph{IEEE Access}, vol.~7, pp. 143\,608--143\,619, 2019.

\bibitem{sayegh2024enhanced}
H.~R. Sayegh, W.~Dong, and A.~M. Al-madani, ``Enhanced intrusion detection with {LSTM}-based model, feature selection, and {SMOTE} for imbalanced data,'' \emph{Applied Sciences}, vol.~14, no.~2, p. 479, 2024.

\bibitem{zhao2017convolutional}
B.~Zhao, H.~Lu, S.~Chen, J.~Liu, and D.~Wu, ``Convolutional neural networks for time series classification,'' \emph{Journal of Systems Engineering and Electronics}, vol.~28, no.~1, pp. 162--169, 2017.

\bibitem{karim2019multivariate}
F.~Karim, S.~Majumdar, H.~Darabi, and S.~Harford, ``Multivariate lstm-fcns for time series classification,'' \emph{Neural Networks}, vol. 116, pp. 237--245, 2019.

\bibitem{mohammadi2024deep}
N.~Mohammadi~Foumani, L.~Miller, C.~W. Tan, G.~I. Webb, G.~Forestier, and M.~Salehi, ``Deep learning for time series classification and extrinsic regression: A current survey,'' \emph{ACM Computing Surveys}, vol.~56, no.~9, pp. 1--45, 2024.

\bibitem{johnson2019survey}
J.~M. Johnson and T.~M. Khoshgoftaar, ``Survey on deep learning with class imbalance,'' \emph{Journal of big data}, vol.~6, no.~1, pp. 1--54, 2019.

\bibitem{wang2023fraud}
X.~Wang, Z.~Liu, J.~Liu, and J.~Liu, ``Fraud detection on multi-relation graphs via imbalanced and interactive learning,'' \emph{Information Sciences}, vol. 642, p. 119153, 2023.

\bibitem{ghosh2024class}
K.~Ghosh, C.~Bellinger, R.~Corizzo, P.~Branco, B.~Krawczyk, and N.~Japkowicz, ``The class imbalance problem in deep learning,'' \emph{Machine Learning}, vol. 113, no.~7, pp. 4845--4901, 2024.

\bibitem{he2009learning}
H.~He and E.~A. Garcia, ``Learning from imbalanced data,'' \emph{IEEE Transactions on Knowledge and Data Engineering}, vol.~21, no.~9, pp. 1263--1284, 2009.

\bibitem{chawla2002smote}
N.~V. Chawla, K.~W. Bowyer, L.~O. Hall, and W.~P. Kegelmeyer, ``{SMOTE}: synthetic minority over-sampling technique,'' \emph{Journal of Artificial Intelligence Research}, vol.~16, pp. 321--357, 2002.

\bibitem{he2008adasyn}
H.~He, Y.~Bai, E.~A. Garcia, and S.~Li, ``{ADASYN}: Adaptive synthetic sampling approach for imbalanced learning,'' in \emph{2008 IEEE International Joint Conference on Neural Networks (IEEE World Congress on Computational Intelligence)}, 2008, pp. 1322--1328.

\bibitem{fernandez2018smote}
A.~Fern{\'a}ndez, S.~Garcia, F.~Herrera, and N.~V. Chawla, ``{SMOTE} for learning from imbalanced data: Progress and challenges, marking the 15-year anniversary,'' \emph{Journal of Artificial Intelligence Research}, vol.~61, pp. 863--905, 2018.

\bibitem{ijcai2021p631}
\BIBentryALTinterwordspacing
Q.~Wen, L.~Sun, F.~Yang, X.~Song, J.~Gao, X.~Wang, and H.~Xu, ``Time series data augmentation for deep learning: A survey,'' in \emph{Proceedings of the Thirtieth International Joint Conference on Artificial Intelligence, {IJCAI-21}}, 8 2021, pp. 4653--4660, survey Track. [Online]. Available: \url{https://doi.org/10.24963/ijcai.2021/631}
\BIBentrySTDinterwordspacing

\bibitem{zhao2022t}
P.~Zhao, C.~Luo, B.~Qiao, L.~Wang, S.~Rajmohan, Q.~Lin, and D.~Zhang, ``{T-SMOTE}: Temporal-oriented synthetic minority oversampling technique for imbalanced time series classification.'' in \emph{IJCAI}, 2022, pp. 2406--2412.

\bibitem{poli2008field}
R.~Poli, W.~B. Langdon, N.~F. McPhee, and J.~R. Koza, ``A field guide to genetic programming,'' 2008.

\bibitem{langdon2013foundations}
W.~B. Langdon and R.~Poli, \emph{Foundations of genetic programming}.\hskip 1em plus 0.5em minus 0.4em\relax Springer Science \& Business Media, 2013.

\bibitem{ahvanooey2019survey}
M.~T. Ahvanooey, Q.~Li, M.~Wu, and S.~Wang, ``A survey of genetic programming and its applications,'' \emph{KSII Transactions on Internet and Information Systems (TIIS)}, vol.~13, no.~4, pp. 1765--1794, 2019.

\bibitem{10793073}
W.~Pei, Y.~Cui, B.~Xue, M.~Zhang, J.~Zhang, Y.~Hou, G.~Zou, and Z.~Qiang, ``{DG-SMOTE}: A distance-angle-based genetic synthetic minority over-sampling technique for unbalanced data learning,'' \emph{IEEE Transactions on Evolutionary Computation}, 2024, \doi{10.1109/TEVC.2024.3515485}.

\bibitem{araf2024cost}
I.~Araf, A.~Idri, and I.~Chairi, ``Cost-sensitive learning for imbalanced medical data: a review,'' \emph{Artificial Intelligence Review}, vol.~57, no.~4, p.~80, 2024.

\bibitem{10068793}
W.~Pei, B.~Xue, M.~Zhang, L.~Shang, X.~Yao, and Q.~Zhang, ``A survey on unbalanced classification: How can evolutionary computation help?'' \emph{IEEE Transactions on Evolutionary Computation}, vol.~28, no.~2, pp. 353--373, 2024.

\bibitem{mohammed2020machine}
R.~Mohammed, J.~Rawashdeh, and M.~Abdullah, ``Machine learning with oversampling and undersampling techniques: overview study and experimental results,'' in \emph{2020 11th international conference on information and communication systems (ICICS)}.\hskip 1em plus 0.5em minus 0.4em\relax IEEE, 2020, pp. 243--248.

\bibitem{bej2021loras}
S.~Bej, N.~Davtyan, M.~Wolfien, M.~Nassar, and O.~Wolkenhauer, ``{LoRAS}: An oversampling approach for imbalanced datasets,'' \emph{Machine Learning}, vol. 110, pp. 279--301, 2021.

\bibitem{pei2023survey}
W.~Pei, B.~Xue, M.~Zhang, L.~Shang, X.~Yao, and Q.~Zhang, ``A survey on unbalanced classification: How can evolutionary computation help?'' \emph{IEEE Transactions on Evolutionary Computation}, vol.~28, no.~2, pp. 353--373, 2024.

\bibitem{elkan2001foundations}
C.~Elkan, ``The foundations of cost-sensitive learning,'' in \emph{International joint conference on artificial intelligence}, vol.~17, no.~1.\hskip 1em plus 0.5em minus 0.4em\relax Lawrence Erlbaum Associates Ltd, 2001, pp. 973--978.

\bibitem{geng2019cost}
Y.~Geng and X.~Luo, ``Cost-sensitive convolutional neural networks for imbalanced time series classification,'' \emph{Intelligent Data Analysis}, vol.~23, no.~2, pp. 357--370, 2019.

\bibitem{gao2020robusttad}
J.~Gao, X.~Song, Q.~Wen, P.~Wang, L.~Sun, and H.~Xu, ``{Robusttad}: Robust time series anomaly detection via decomposition and convolutional neural networks,'' \emph{arXiv preprint arXiv:2002.09545}, 2020.

\bibitem{oksuz2020imbalance}
K.~Oksuz, B.~C. Cam, S.~Kalkan, and E.~Akbas, ``Imbalance problems in object detection: A review,'' \emph{IEEE Transactions on Pattern Analysis and Machine Intelligence}, vol.~43, no.~10, pp. 3388--3415, 2020.

\bibitem{han2005borderline}
H.~Han, W.-Y. Wang, and B.-H. Mao, ``{Borderline-SMOTE}: a new over-sampling method in imbalanced data sets learning,'' in \emph{International Conference on Intelligent Computing}.\hskip 1em plus 0.5em minus 0.4em\relax Springer, 2005, pp. 878--887.

\bibitem{zhang2021surrogate}
F.~Zhang, Y.~Mei, S.~Nguyen, M.~Zhang, and K.~C. Tan, ``Surrogate-assisted evolutionary multitask genetic programming for dynamic flexible job shop scheduling,'' \emph{IEEE Transactions on Evolutionary Computation}, vol.~25, no.~4, pp. 651--665, 2021.

\bibitem{wang2025semantics}
C.~Wang, Q.~Chen, B.~Xue, and M.~Zhang, ``Semantics-guided multi-task genetic programming for multi-output regression,'' \emph{Pattern Recognition}, vol. 161, p. 111289, 2025.

\bibitem{fonseca2023comparing}
A.~Fonseca and D.~Po{\c{c}}as, ``Comparing the expressive power of strongly-typed and grammar-guided genetic programming,'' in \emph{Proceedings of the Genetic and Evolutionary Computation Conference}, 2023, pp. 1100--1108.

\bibitem{montana1995strongly}
D.~J. Montana, ``Strongly typed genetic programming,'' \emph{Evolutionary Computation}, vol.~3, no.~2, pp. 199--230, 1995.

\bibitem{abanda2019review}
A.~Abanda, U.~Mori, and J.~A. Lozano, ``A review on distance based time series classification,'' \emph{Data Mining and Knowledge Discovery}, vol.~33, no.~2, pp. 378--412, 2019.

\bibitem{kate2016using}
R.~J. Kate, ``Using dynamic time warping distances as features for improved time series classification,'' \emph{Data Mining and Knowledge Discovery}, vol.~30, pp. 283--312, 2016.

\bibitem{sundararajan2024discrete}
D.~D. Sundararajan, ``Discrete fourier transform,'' in \emph{Digital Signal Processing: An Introduction}.\hskip 1em plus 0.5em minus 0.4em\relax Springer, 2024, pp. 67--106.

\bibitem{1705539}
C.-S. Yu, ``A discrete fourier transform-based adaptive mimic phasor estimator for distance relaying applications,'' \emph{IEEE Transactions on Power Delivery}, vol.~21, no.~4, pp. 1836--1846, 2006.

\bibitem{yuan2024diffusionts}
\BIBentryALTinterwordspacing
X.~Yuan and Y.~Qiao, ``Diffusion-{TS}: Interpretable diffusion for general time series generation,'' in \emph{The Twelfth International Conference on Learning Representations}, 2024. [Online]. Available: \url{https://openreview.net/forum?id=4h1apFjO99}
\BIBentrySTDinterwordspacing

\bibitem{10120936}
B.~Al-Helali, Q.~Chen, B.~Xue, and M.~Zhang, ``Multitree genetic programming with feature-based transfer learning for symbolic regression on incomplete data,'' \emph{IEEE Transactions on Cybernetics}, vol.~54, no.~7, pp. 4014--4027, 2024.

\bibitem{fortin2012deap}
F.-A. Fortin, F.-M. De~Rainville, M.-A.~G. Gardner, M.~Parizeau, and C.~Gagn{\'e}, ``{DEAP}: Evolutionary algorithms made easy,'' \emph{The Journal of Machine Learning Research}, vol.~13, no.~1, pp. 2171--2175, 2012.

\bibitem{lemaavztre2017imbalanced}
G.~Lema{\~A}{\v{Z}}tre, F.~Nogueira, and C.~K. Aridas, ``Imbalanced-learn: A python toolbox to tackle the curse of imbalanced datasets in machine learning,'' \emph{Journal of Machine Learning Research}, vol.~18, no.~17, pp. 1--5, 2017.

\bibitem{10.1007/978-3-319-08010-9_33}
Y.~Zheng, Q.~Liu, E.~Chen, Y.~Ge, and J.~L. Zhao, ``Time series classification using multi-channels deep convolutional neural networks,'' in \emph{Web-Age Information Management}, F.~Li, G.~Li, S.-w. Hwang, B.~Yao, and Z.~Zhang, Eds.\hskip 1em plus 0.5em minus 0.4em\relax Cham: Springer International Publishing, 2014, pp. 298--310.

\bibitem{rigatti2017random}
S.~J. Rigatti, ``Random forest,'' \emph{Journal of Insurance Medicine}, vol.~47, no.~1, pp. 31--39, 2017.

\bibitem{liu2019model}
C.-L. Liu and P.-Y. Hsieh, ``Model-based synthetic sampling for imbalanced data,'' \emph{IEEE Transactions on Knowledge and Data Engineering}, vol.~32, no.~8, pp. 1543--1556, 2019.

\bibitem{yin2020novel}
J.~Yin, C.~Gan, K.~Zhao, X.~Lin, Z.~Quan, and Z.-J. Wang, ``A novel model for imbalanced data classification,'' in \emph{Proceedings of the AAAI Conference on Artificial Intelligence}, vol.~34, no.~04, 2020, pp. 6680--6687.

\bibitem{fawcett2006introduction}
T.~Fawcett, ``An introduction to {ROC} analysis,'' \emph{Pattern Recognition Letters}, vol.~27, no.~8, pp. 861--874, 2006.

\bibitem{lee2018proper}
S.~Lee and D.~K. Lee, ``What is the proper way to apply the multiple comparison test?'' \emph{Korean journal of anesthesiology}, vol.~71, no.~5, pp. 353--360, 2018.

\bibitem{demvsar2006statistical}
J.~Dem{\v{s}}ar, ``Statistical comparisons of classifiers over multiple data sets,'' \emph{Journal of Machine learning research}, vol.~7, no. Jan, pp. 1--30, 2006.

\end{thebibliography}

\end{document}